%% file: main.tex
\documentclass{article}
\PassOptionsToPackage{numbers,sort&compress}{natbib}
\usepackage{arxivpaper}

\usepackage[hyphens]{url}
\usepackage{graphicx}
\usepackage{booktabs}
\usepackage{multirow}
\usepackage{tikz}
\usepackage{colortbl}
\usepackage{fontawesome5}
\usepackage{microtype}
\usepackage[hidelinks,
  pdftitle={Lethe: How Hard Is It to Forget? A Benchmark for Federated Unlearning in Medical Imaging},
  pdfauthor={Shengchao Chen and Ting Shu}]{hyperref}
\definecolor{shadebest}{HTML}{C8E6C9}
\definecolor{shade2nd}{HTML}{E8F5E9}
\definecolor{linkpink}{HTML}{D81B60}
\renewcommand{\arraystretch}{1.1}

\newcommand{\HBr}{0.4em}
\DeclareRobustCommand{\HBfull}{\tikz[baseline=-0.55ex]{\fill (0,0) circle (\HBr);}}
\DeclareRobustCommand{\HBhalf}{\tikz[baseline=-0.55ex]{\draw[line width=0.4pt] (0,0) circle (\HBr);\fill (0,\HBr) arc (90:270:\HBr) -- cycle;}}
\DeclareRobustCommand{\HBnone}{\tikz[baseline=-0.55ex]{\draw[line width=0.4pt] (0,0) circle (\HBr);}}

\title{Lethe: How Hard Is It to Forget? A Benchmark for Federated Unlearning in Medical Imaging}

\newcommand{\blfootnote}[1]{%
  \begingroup
  \renewcommand{\thefootnote}{}%
  \footnote{#1}%
  \addtocounter{footnote}{-1}%
  \endgroup
}

\author{%
  Shengchao Chen \\
  University of Technology Sydney \\
  \And
  Ting Shu \\
  Shenzhen University \\
}

\begin{document}
\maketitle
\blfootnote{Correspondence to \texttt{shengchao.chen.uts@gmail.com}.}

\begin{abstract}
Federated learning enables medical-imaging models to be trained across hospitals, and privacy law, most explicitly the GDPR ``right to be forgotten'', turns removing a hospital's, a class's, or a patient's influence from such a model into a federated unlearning problem. This need is most acute in medicine, where patients withdraw consent and hospitals leave collaborations. Yet nearly all unlearning evidence comes from natural images, whose heterogeneity and task structure differ sharply from clinical data, so it is unclear whether existing methods transfer, and no shared protocol covers clinical data. We present \textsc{Lethe}, a benchmark for federated unlearning in medical imaging. It evaluates twelve methods across eight task families, from classification and segmentation to denoising, cross-modality synthesis, and vision-language question answering, at three forgetting granularities and against a retrained gold standard on utility, privacy, and cost. The central result is that what separates methods is the difficulty of the forgetting request, not the method itself. The easy removals that dominate the literature leave the methods that preserve utility indistinguishable, while only hard ones separate them. More striking, on the many medical tasks that generalize across sites, forgetting a client barely changes task performance, leaving residual membership as the signal that must be erased.
\end{abstract}

\begin{center}
\faGithub\hspace{0.4em}\textcolor{linkpink}{\url{https://github.com/shengchaochen82/Lethe}}
\end{center}

\section{Introduction}
Deep learning underpins much of medical imaging~\citep{chen2023interpretable}, yet the scans, slides, and records behind these models are sensitive and siloed across hospitals that rarely pool them, for privacy and policy. Federated learning (FL) addresses this by training a shared model while each site's data never leaves it~\citep{fedavg}, and is increasingly used for clinical imaging~\citep{flamby}. Privacy law also grants a \emph{right to be forgotten}, most explicitly in the GDPR: once a model is trained, a hospital may leave, a patient may withdraw consent, or a label may be retired, and the model's dependence on that data must be removed.

Retraining from scratch removes that dependence exactly, but across many hospitals it is prohibitive. This cost has motivated \emph{federated unlearning} (FU), a growing family of methods that approximate the retrained model more cheaply, from historical-update calibration and gradient ascent to recent distillation and projection schemes~\citep{federaser,halimi,fusurvey}. Almost all of them, however, are designed and validated on natural images, where machine unlearning has largely developed, so a hospital federation must choose among them with no evidence that any behaves well on clinical data.

Medical imaging is where the right to be forgotten most directly applies. Clinical data is partitioned by hospital, spans modalities and tasks far from natural-image classification, and faces stricter privacy scrutiny, so conclusions from natural images need not transfer to clinical data. Evaluation here is nonetheless fragmented. General unlearning benchmarks cover only natural images, centralized~\citep{deepunlearn} and federated~\citep{fusurvey}, and the few medical studies each test a single method on a small number of datasets under one heterogeneity setting~\citep{fcu,maverick}, with no shared protocol across the requests, tasks, and architectures, leaving no basis for determining which method is reliable.

To address this gap, we present \textsc{Lethe}, named for the mythological river of forgetting, a benchmark for federated unlearning in medical imaging. It brings together twelve unlearning methods, eight task families spanning classification, segmentation, denoising, 3D classification, MRI synthesis, registration, lesion localization, and medical visual question answering, three kinds of forgetting request (a departing hospital, a retired label, a withdrawn patient), varied heterogeneity, and a seven-metric panel under one shared protocol. One result reframes the rest. On the easy requests that dominate the literature no method that preserves utility can be distinguished from another, and what decides whether a benchmark separates them at all is the difficulty of the forgetting request, not the choice of method. Our contributions are as follows.

\begin{itemize}
  \item We present \textsc{Lethe}, a benchmark for federated unlearning in medical imaging, pairing representative methods from seven mechanism families with tasks spanning eight task families and sixteen datasets, under standardized splits, a common protocol, and a seven-metric panel.
  \item We find that forgetting difficulty, not the choice of method, decides whether methods separate, and build a difficulty-aware protocol on hard requests such as class-level and sole-class removal.
  \item We show that on the many clinical tasks that generalize across sites, forgetting a client barely changes the task metric, leaving membership rather than task contribution as the signal an unlearner must erase.
\end{itemize}

\section{Related Work}

\paragraph{Federated Medical Imaging.} Federated learning is an established paradigm for medical imaging, training a shared model across hospitals while patient data stays local~\citep{fedavg}. The setting is cross-silo rather than cross-device: a small number of sites with abundant but private data, whose scanners, acquisition protocols, and patient populations make the data non-IID by construction. FL now spans clinical tasks, from classification and lesion detection to segmentation and reconstruction, with cross-silo benchmarks standardizing these tasks and hospital splits~\citep{flamby}. A recent line adapts large vision-language models to federated medical imaging under severe label and modality skew~\citep{chen2025restyled,feng2025taming,feng2026visual}, within a wider move toward federated foundation models over heterogeneous, device-held data~\citep{chen2023prompt,chen2023federated,chen2025federated,chen2026fedal}. Almost all of this effort targets \emph{training} accurate models. The complementary problem of \emph{removing} a site's or a patient's contribution afterwards has drawn far less attention. The few medical federated-unlearning studies each validate a single method on a narrow setup~\citep{fcu,maverick}, with no common evaluation protocol across the tasks and partitions that define clinical FL.

\paragraph{Federated Unlearning.} Federated unlearning methods differ mainly in mechanism, from historical-update calibration~\citep{federaser} and projected gradient ascent~\citep{halimi} to pruning~\citep{classprune}, distillation~\citep{fedquit}, and the noise~\citep{feddni}, overwriting~\citep{fused}, and conflict-aware projection~\citep{fedcare} schemes, all surveyed by \citet{fusurvey}. Evaluation has not kept pace: general benchmarks cover only centralized natural images~\citep{deepunlearn}, federated ones use one or two natural-image datasets under a single heterogeneity setting, and privacy is judged by a population membership-inference attack~\citep{shokri_mia,nasr_fl} rather than the per-sample attacks standard in centralized unlearning~\citep{carlini_lira,ulira}. How these methods behave on medical data, across varied forgetting requests, and under a calibrated per-sample attack is therefore not systematically evaluated. \textsc{Lethe} supplies that protocol, and \textbf{Table~\ref{tab:compare}} summarizes how it differs from recent evaluations.

\begin{table}[tbh]
\centering
\caption{Lethe versus related evaluations. \HBfull~full support, \HBhalf~partial, \HBnone~absent.}
\label{tab:compare}
\setlength{\tabcolsep}{2.5pt}\renewcommand{\arraystretch}{1.2}
\resizebox{\textwidth}{!}{%
\begin{tabular}{@{}lcccccccccc@{}}
\toprule
& \multicolumn{3}{c}{\textbf{Scope}} & \multicolumn{5}{c}{\textbf{Coverage}} & \multicolumn{2}{c}{\textbf{Realism}} \\
\cmidrule(lr){2-4}\cmidrule(lr){5-9}\cmidrule(lr){10-11}
Benchmark & Medical & Federated & Unlearning & \#\,Tasks & \#\,Data & \#\,Methods & Granularity & Heterog. & Real clinical & Metric panel \\
\midrule
FLamby~\citep{flamby} & \HBfull & \HBfull & \HBnone & 3 & 7 & -- & -- & \HBnone & \HBfull & \HBnone \\
Deep Unlearn~\citep{deepunlearn} & \HBnone & \HBnone & \HBfull & 1 & 5 & 18 & sample & \HBnone & \HBnone & \HBhalf \\
General FU~\citep{fusurvey} & \HBnone & \HBfull & \HBfull & 1 & -- & -- & client & \HBnone & \HBnone & \HBhalf \\
FCU~\citep{fcu} & \HBfull & \HBfull & \HBfull & 1 & 2 & 1 & client & \HBnone & \HBhalf & \HBhalf \\
Maverick~\citep{maverick} & \HBfull & \HBfull & \HBfull & 1 & 3 & 1 & client/class/sample & \HBnone & \HBhalf & \HBhalf \\
\midrule
\textbf{Lethe (ours)} & \HBfull & \HBfull & \HBfull & \textbf{8} & \textbf{16} & \textbf{12} & client/class/sample & \HBfull & \HBfull & \HBfull \\
\bottomrule
\end{tabular}}
\end{table}

\section{The \textsc{Lethe} Benchmark}

\paragraph{Problem Formulation.} A federation of $K$ clients, client $k$ holding a local dataset $D_k$ with $D=\bigcup_{k}D_k$, trains a shared model by FedAvg~\citep{fedavg}. Writing $\mathrm{FA}(S)$ for the model FedAvg returns from data $S$, the pre-unlearn model and the gold reference are
\begin{equation}
M_0=\mathrm{FA}(D),\qquad M^\ast=\mathrm{FA}(D_r),
\label{eq:setup}
\end{equation}
where a forgetting request removes a subset $D_f\subseteq D$ and leaves the retain set $D_r=D\setminus D_f$. The request arrives at one of three granularities,
\begin{equation}
\underbrace{D_f=D_k}_{\text{client}},\qquad
\underbrace{D_f=\{(x,y)\in D:\,y{=}c\}}_{\text{class}},\qquad
\underbrace{D_f\subset D_k}_{\text{sample}}.
\label{eq:granularity}
\end{equation}
The pre-unlearn model $M_0$ depends on every $D_k$, whereas $M^\ast$ is the gold-standard reference every method approximates. An unlearning method is an operator $\mathcal{U}$ producing the unlearned model
\begin{equation}
M_u=\mathcal{U}(M_0,D_f,D_r),
\label{eq:operator}
\end{equation}
and its aim is $M_u\approx M^\ast$ at cost far below recomputing $M^\ast$, using only the information its own formulation assumes, from the forget data alone to stored update history (\textbf{App.~\ref{app:methods}}).
\begin{figure}[tbh]\centering
\includegraphics[width=\textwidth]{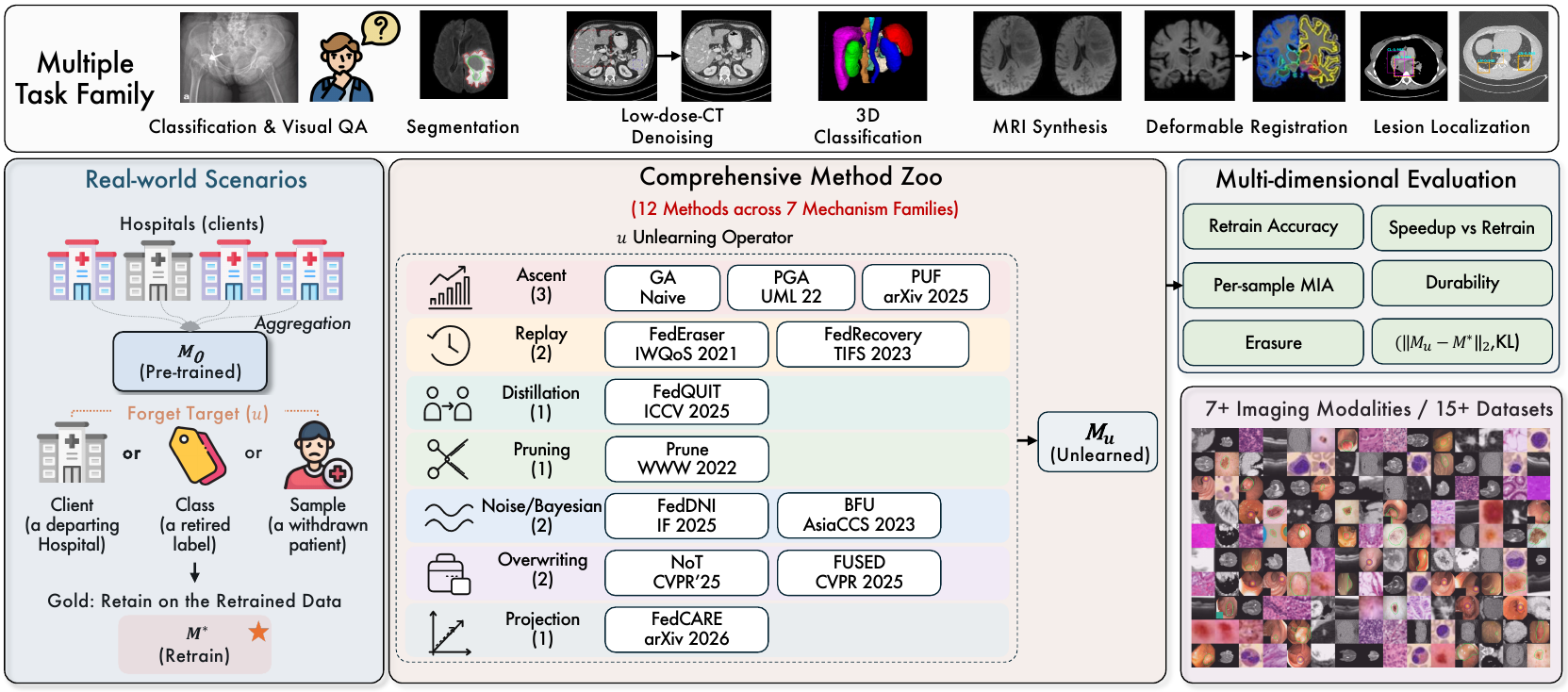}
\caption{Overview of \textsc{Lethe}. A federated model $M_0$ receives a forgetting request $u$ at client, class, or sample granularity, and an unlearning operator maps it to $M_u$, scored against the gold-standard retrain $M^\ast$. The benchmark spans eight task families, twelve methods, and a seven-metric panel covering utility, privacy, backdoor erasure, efficiency, and durability.}
\label{fig:overview}
\end{figure}
\paragraph{Benchmark Overview.} \textsc{Lethe} asks not which method forgets best, but whether current evaluations can distinguish methods at all. This motivates two design axes. First, forgetting \emph{difficulty}: we include hard requests such as class-level removal and sole-class departure, where the leaving client is a class's only holder, since common client-level removal often fails to separate methods. Second, we span tasks with different degrees of cross-site generalization, distinguishing genuine forgetting from vacuous signals caused by task transfer. Each experiment pairs a standardized federated \emph{dataset}, a \emph{method}, and a full-metric \emph{evaluation} against gold retraining. \textbf{Fig.~\ref{fig:overview}} summarizes this pipeline.

\paragraph{Datasets and Tasks.} \textsc{Lethe} spans eight task families (\textbf{Table~\ref{tab:tasks}}) chosen to vary forgetting difficulty and cross-site generalization. Classification covers six datasets, four from MedMNIST~\citep{medmnistv2} (blood, abdominal CT, colon histology, retinal OCT), the harder Kvasir GI-endoscopy set~\citep{kvasir}, and Camelyon17 across five hospitals~\citep{bandi17,wilds}. Six further families reach beyond 2D classification, where a task often transfers across sites: segmentation~\citep{kvasirseg,cvcclinicdb,isic}, low-dose-CT denoising~\citep{mayo2016}, 3D classification, MRI synthesis and deformable registration on IXI~\citep{ixi}, and lesion localization. An eighth, vision-language family covers multimodal medical question answering (VQA-RAD, SLAKE, and PathVQA~\citep{vqarad,slake,pathvqa}). Unless stated otherwise, MedMNIST and Kvasir carry no natural client split, so we partition them across $20$ clients by a Dirichlet($\alpha$) label-skew ($\alpha{=}0.1$)~\citep{dirichletpart}, a severe cross-silo label skew, while Camelyon17 follows its natural five-hospital split.

\begin{table}[t]\centering\small
\caption{The eight task families in \textsc{Lethe}, with their datasets, networks, and utility metrics, all additionally scored on shared privacy, efficiency, and durability.}
\label{tab:tasks}
\input{tables/table_tasks.tex}

\end{table}

\paragraph{Methods and Mechanisms.} \textsc{Lethe} evaluates twelve unlearning methods against the gold \emph{Retrain}, each reported in \textbf{Table~\ref{tab:main}} and grouped by their primary mechanism to span what federated unlearning has explored. \emph{Ascent} methods move the model against the forget loss: gradient ascent (GA), its projected variant PGA~\citep{halimi} that confines each step to an $L_2$ ball around a reference, and PUF~\citep{puf} that negates pseudo-gradients. \emph{Replay} methods reuse stored training history: FedEraser~\citep{federaser} recalibrates saved client updates, and FedRecovery~\citep{fedrecovery} subtracts a weighted sum of gradient residuals with differential-privacy noise. The remaining families each edit the model differently: \emph{distillation} against a virtual teacher (FedQUIT~\citep{fedquit}), class-discriminative \emph{pruning} of the dimensions most attributable to the forget target (Prune~\citep{classprune}), \emph{noise and Bayesian} erasure by diffusive injection (FedDNI~\citep{feddni}) or variational self-sharing (BFU~\citep{bfu}), \emph{overwriting} by weight negation (NoT~\citep{not}) or sparse adapters (FUSED~\citep{fused}), and conflict-aware \emph{projection} that removes the forget gradient while preserving the retained loss (FedCARE~\citep{fedcare}).
\paragraph{Granularities.} We keep all three granularities because they differ sharply in difficulty, unlike the client-level-only evaluations of most prior work. We also remove several clients at once and study \emph{sequential} departure, checking that erasure does not degrade as requests accumulate.

\paragraph{Metrics.} A single metric cannot certify forgetting. Forget accuracy alone is driven to zero by any model that simply degrades, so we score every unlearned model against the gold $M^\ast$ across a panel and read each axis relative to $M^\ast$, not to zero. Let $A(M,D)$ be the accuracy of $M$ on the evaluation data of partition $D$, and $D_t$ the global test set. \emph{Utility} is the retain and test accuracy $A(M_u,D_r)$ and $A(M_u,D_t)$, which must stay close to $M^\ast$, while the forget accuracy $A(M_u,D_f)$ is reported but never rewarded on its own. \emph{Privacy} is a per-sample membership-inference attack~\citep{carlini_lira,ulira} that can expose residual leakage even when forget accuracy already matches $M^\ast$. \emph{Backdoor erasure} tests whether a planted trigger is removed, \emph{cost} is the speedup over retraining, and \emph{durability} is whether forgetting survives fine-tuning,
\begin{equation}
\Delta_2=\lVert \theta_u-\theta^\ast\rVert_2,\quad
s=\frac{T_{M^\ast}}{T_{M_u}},\quad
R_k=A\!\big(M_u^{(k)},D_f\big),
\label{eq:metrics}
\end{equation}
where $\theta$ denotes parameters and $M_u^{(k)}$ is $M_u$ after $k$ fine-tuning epochs on $D_f$, so a durable erasure keeps $R_k$ near $M^\ast$, the retrain gap $\Delta_2$ and $\mathrm{KL}(M_u\Vert M^\ast)$ measure closeness to gold, and $s$ its cost.

\section{Experiments and Results}\label{sec:findings}
\subsection{Federated Medical Image Classification}

\paragraph{Setup.} Classification is the most common computer-aided-diagnosis task. Each dataset is split across 20 clients by a Dirichlet label skew ($\alpha{=}0.1$, varied over $\{0.1,0.5,1.0\}$ on MedMNIST). Models are a compact two-block CNN and a from-scratch ResNet-18~\citep{resnet}, trained with FedAvg~\citep{fedavg} and SGD ($150$ rounds, learning rate $0.005$, batch $64$, Camelyon17 $50$ rounds at batch $256$). Each pre-trained model is reused across methods, unlearning for five epochs then five recovery rounds on retained clients (per-method hyperparameters in \textbf{App.~\ref{app:methods}}). The forget target is client~$0$, class~$2$, or a random $10\%$ of each client's samples, scored against the gold $M^\ast$ over three seeds.
\begin{table}[tbh]\centering\small
\caption{Client-level forgetting across all datasets, with forget (F) and retain (R) accuracy for every method. Retrain$^*$ is the gold standard and \emph{Full model}$^\dagger$ the pre-unlearn reference ($F_0$). Green marks the two retain values nearest gold (darker = closest), here and in later tables. The full heterogeneity range is in \textbf{App.~\ref{app:hetero}}.}
\label{tab:main}
\setlength{\tabcolsep}{2pt}
\resizebox{\textwidth}{!}{\input{tables/table1_main.tex}}

\end{table}

\begin{figure}[tbh]\centering
\includegraphics[width=.62\textwidth]{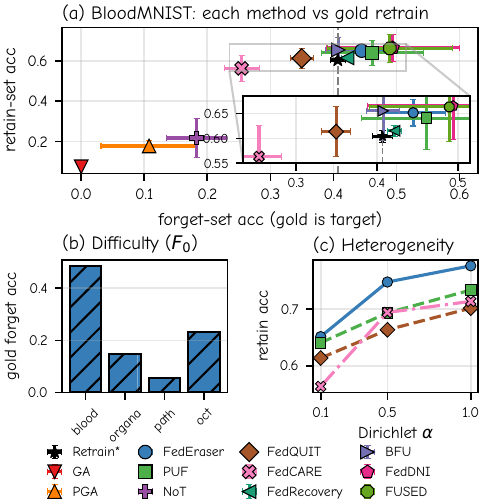}
\caption{(a) Forget vs retain accuracy on BloodMNIST (easiest request), gentle cluster magnified in the inset. (b) Pre-unlearn forget accuracy $F_0$ per dataset. (c) Retain accuracy vs Dirichlet $\alpha$ on BloodMNIST, four gentle methods.}\label{fig:results}
\end{figure}

\paragraph{Difficulty Governs Separability.} \textbf{Fig.~\ref{fig:results}(a)} shows forget and retain accuracy on the easiest client-level request, where the gentle methods collapse into one overlapping band, so no ranking is stable. What a benchmark can resolve is set by difficulty, read from the pre-unlearn forget accuracy $F_0$ (\textbf{Fig.~\ref{fig:results}(b)}), which falls from $0.49$ on BloodMNIST to $0.05$ on PathMNIST, where the departing client's classes are covered by others and every gold-tracking method matches retrain (\textbf{Table~\ref{tab:main}}). Harder requests confirm the mechanism. Class-level removal separates methods (\textbf{Table~\ref{tab:classlevel}}), with pruning and FedQUIT holding retain near gold while gradient ascent collapses it, though even accuracy-matching methods can leave the forget class clustered, erasing the label but not the representation. \textbf{App.~\ref{app:repr}} extends this to a sole-holder protocol, testing whether methods still separate when the forget client is a class's only holder.

\begin{table}[t]\centering\small
\caption{Class-level retain accuracy per method and dataset (the near-zero forget-class accuracy is omitted).}\label{tab:classlevel}
\input{tables/table3_classlevel.tex}

\end{table}

\paragraph{Heterogeneity, Scale, and Durability.} \textbf{Fig.~\ref{fig:results}(c)} shows that varying $\alpha$ shifts retain accuracy but leaves the gentle methods clustered, while only ascent methods deteriorate. Heterogeneity therefore determines the available accuracy rather than the preferred method. \textbf{App.~\ref{app:scale}} further enlarges the federation at fixed $\alpha$: gold forget accuracy remains high across federation sizes, whereas fewer samples per client reduce retain accuracy and destabilize FedCARE, ruling out size alone as the cause. Durability instead depends on the removal target. Backdoors do not return and sequential departures do not compound, whereas class-level forgetting is fragile: only three relearning epochs restore the class even after gold retraining, because removal resets the head but leaves the underlying features intact.

\paragraph{Privacy and Robustness.} Privacy requires a per-sample attack. The population attack stays near chance, whereas the retrain-calibrated LiRA~\citep{carlini_lira} (\textbf{Table~\ref{tab:mia}}) exposes residual membership on BloodMNIST for FedEraser, PUF, FedRecovery, and FedDNI while FedQUIT and FedCARE match retrain. \textbf{App.~\ref{app:privacy}} repeats this with the stronger U-LiRA~\citep{ulira}, isolating the residual signal at a low false-positive rate. Distance to gold agrees, gentle methods staying within $\Delta_2\approx0.2$--$0.5$ at near-zero KL while ascent and heavy-noise methods drift far (\textbf{App.~\ref{app:retraingap}}). \textbf{App.~\ref{app:robustness}} confirms this under sample-level forgetting, two-client removal, and a from-scratch ResNet-18.

\begin{table}[t]\centering
\caption{Retrain-calibrated per-sample MIA on MedMNIST (ideal $0.5$). Dashes mark unavailable BFU runs.}\label{tab:mia}
\input{tables/table6_mia.tex}

\end{table}

\subsection{Federated Medical Image Segmentation}
\paragraph{Setup.} Segmentation, which delineates organs and lesions, spans three datasets and two modalities (Kvasir-SEG~\citep{kvasirseg}, CVC-ClinicDB~\citep{cvcclinicdb}, ISIC-2016~\citep{isic}), each learned by a U-Net~\citep{unet} under a Dice-plus-cross-entropy loss. Clients follow a Dirichlet quantity skew ($\alpha{=}0.5$, heterogeneity in data volume rather than label mix), running $80$ rounds of two local epochs at learning rate $0.02$ and batch $8$, over three seeds.

\begin{figure}[tbh]\centering
\includegraphics[width=\textwidth]{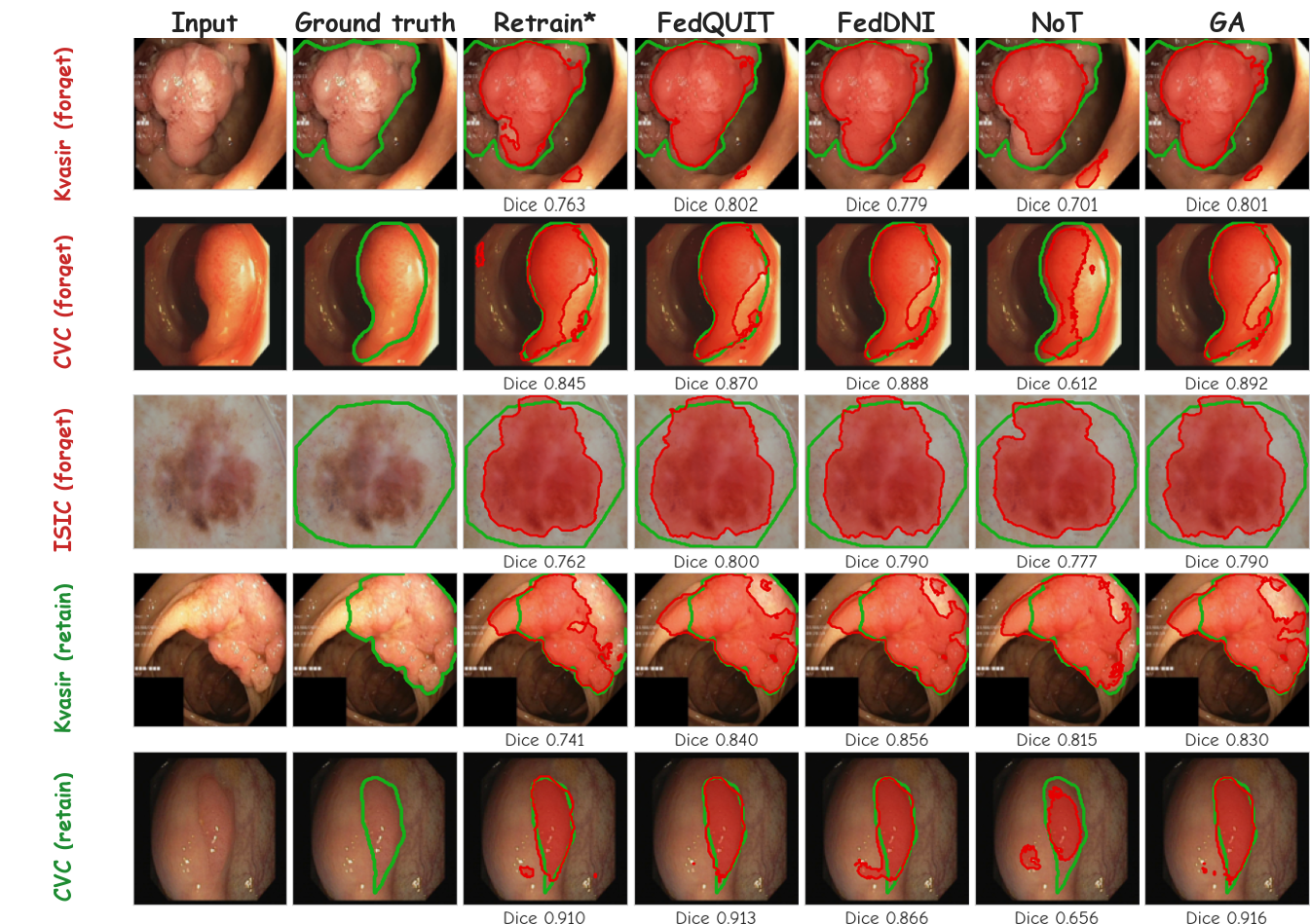}
\caption{Segmentation case study across three datasets, showing forget (red) and retained (green) clients against the input, ground truth, the gold retrain, and the unlearning methods, with per-image Dice. Full studies in \textbf{App.~\ref{app:casestudies}}.}
\label{fig:segteaser}
\end{figure}

\begin{table}[tbh]\centering
\caption{Segmentation unlearning, with forget (F) and retain (R) Dice and a retrain-calibrated membership MIA (-- for the gold). $F_0$ is the pre-unlearn forget Dice.}\label{tab:seg}
\resizebox{\textwidth}{!}{\input{tables/table14_segmentation.tex}}

\end{table}
\paragraph{Results.} \textbf{Table~\ref{tab:seg}} reports forget and retain Dice for eleven methods. The gold retrain, which was never trained on the forget client, still segments its data within a few percentage points of every method (\textbf{Fig.~\ref{fig:segteaser}}). A U-Net generalizes across sites, so dropping one client barely changes what it segments and the forget metric is nearly vacuous. What little separates methods is again membership, where FedRecovery leaks most. \textbf{App.~\ref{app:robustness}} reports sample-level forgetting, repeating the classification dataset-dependence.

\subsection{Vision-Language: Medical VQA}
\begin{table}[tbh]\centering\small
\caption{Client-level medical VQA, closed-set classification (top) and open-ended generation (bottom), with forget (F) and retain (R) accuracy, a MIA, and retain $R_v$ under a stronger ResNet-18 and DistilBERT encoder.}\label{tab:vqa}
\resizebox{\textwidth}{!}{\input{tables/table_vqa_combined.tex}}

\end{table}
\paragraph{Setup.} Medical VQA pairs an image with a free-text question. We use three closed-set VQA datasets, VQA-RAD~\citep{vqarad} and SLAKE~\citep{slake} (radiology) and PathVQA~\citep{pathvqa} (pathology), each cast as classification over its thirty most frequent answers and learned by a small dual-encoder fusing a CNN image branch and a word-embedding question branch. The protocol runs eight to twelve clients per dataset and eleven methods over three seeds, and we also evaluate a stronger ResNet-18 and DistilBERT~\citep{distilbert} encoder on all three.
\paragraph{Results.} \textbf{Table~\ref{tab:vqa}} reports client-level forgetting on three VQA datasets. On VQA-RAD, the forget client's accuracy falls from $0.49$ to the gold retrain's $0.24$, with all methods approaching gold ($0.16$--$0.24$), retain accuracy remaining near $0.48$, and MIA staying near chance. PathVQA generalizes sufficiently well that little remains to forget (forget $0.50$ vs.\ gold $0.51$), whereas the harder SLAKE retains a small gap ($0.21$ vs.\ $0.16$) that only ascent methods reduce. These results indicate that erasability is governed by how strongly the forgotten client contributes beyond cross-client task generalization. The same pattern persists with the stronger encoder despite dataset-dependent shifts in absolute retain accuracy, as shown by the $R_v$ columns. In the open-ended setting, scored by token accuracy in the lower block, gentle methods also match the gold retrain across all three datasets with both the small encoder and the stronger VLM. \textbf{App.~\ref{app:vqa}} verifies this conclusion at finer granularities and under a stronger per-sample membership attack.

\subsection{Further Task Families}

\begin{figure}[tbh]\centering
\includegraphics[width=\textwidth]{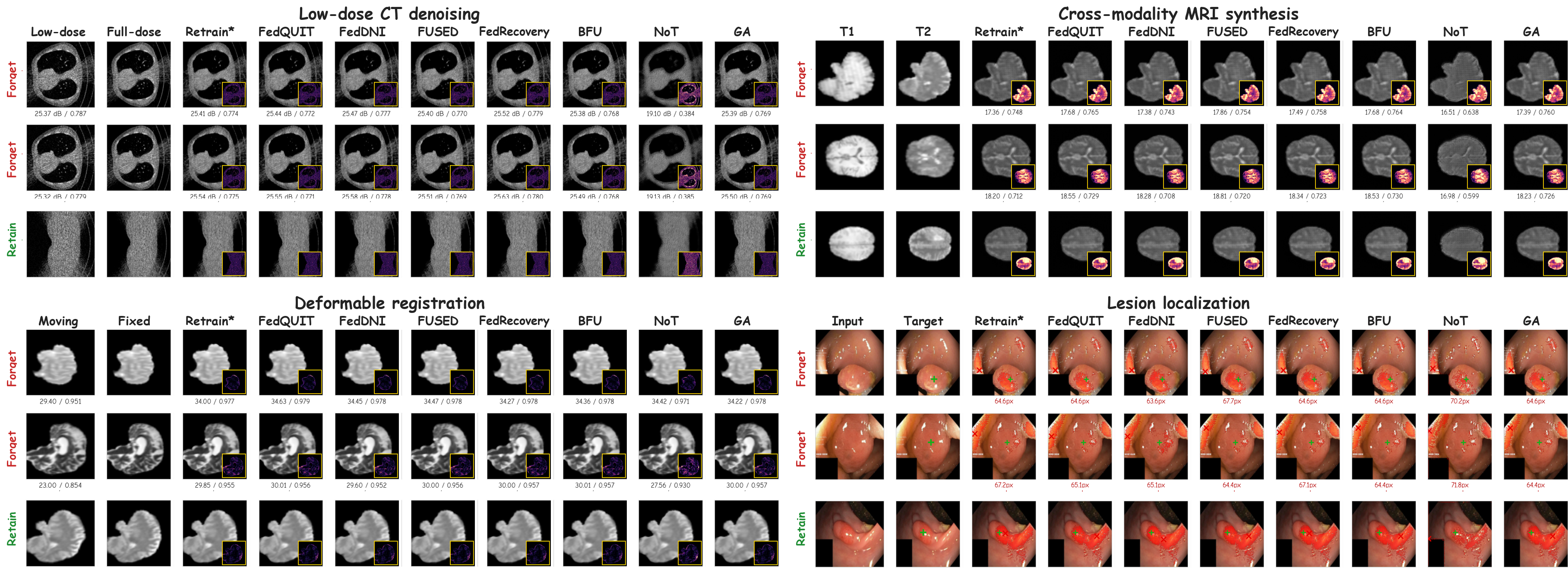}
\caption{Case studies on four further task families, each showing two forget (red) and one retained (green) case against the task input, target, the gold retrain, and the unlearning methods. Insets are per-pixel error maps where present, and the localization panel marks each method's predicted centroid ($\times$) against ground truth ($+$). Full studies for every method are in \textbf{App.~\ref{app:casestudies}}.}
\label{fig:furthercase}
\end{figure}

\paragraph{Setup.} We evaluate five remaining clinical task families at the client level over three seeds with all ten applicable methods. Low-dose CT denoising uses AAPM-Mayo-2016~\citep{mayo2016}, ten patient-clients, a U-Net, and PSNR. 3D classification uses MedMNIST3D~\citep{medmnistv2}, $28^3$ inputs, and a small 3D CNN trained for $60$ rounds with batch size $32$ and learning rate $0.005$. Cross-modality MRI synthesis and deformable registration use three-hospital IXI~\citep{ixi}, $80$ rounds, batch size $8$, learning rate $0.02$, and PSNR/SSIM. Registration applies a VoxelMorph-style deformation field~\citep{voxelmorph} to warp the moving image. Lesion localization reuses the three segmentation datasets~\citep{kvasirseg,cvcclinicdb,isic}, turning each mask into a Gaussian-heatmap centroid.
\begin{table}[tbh]\centering
\caption{Low-dose-CT denoising, with forget, retain, and test PSNR (dB) and a MIA ($F_0$: pre-unlearn forget).}\label{tab:denoise}
\input{tables/table_denoise.tex}

\end{table}
\paragraph{Results.} \textbf{Tables~\ref{tab:denoise}--\ref{tab:synthreg}} report all five. Each repeats the pattern. Forgetting a client barely changes the metric and every method clusters at the gold retrain, leaving membership rather than the task metric as the erasable signal. The two exceptions are the hard OrganMNIST3D, where high difficulty splits methods in a 3D architecture, and NoT, whose weight negation collapses the denoiser (\textbf{Fig.~\ref{fig:furthercase}}). Across all eight task families, difficulty and cross-site generalization, not the unlearning algorithm, govern what can be distinguished.

\begin{table}[tbh]\centering
\caption{3D volumetric classification on MedMNIST3D (accuracy) and lesion localization by Gaussian-heatmap centroid (hit-rate), with forget (F) and retain (R) scores.}\label{tab:cls3ddetect}
\resizebox{\textwidth}{!}{\input{tables/table_cls3d_detect.tex}}

\end{table}

\begin{table}[tbh]\centering
\caption{Cross-modality MRI synthesis (IXI T1$\to$T2) and deformable registration (VoxelMorph-style), with forget and retain PSNR (dB) and SSIM against the target.}\label{tab:synthreg}
\resizebox{\textwidth}{!}{\input{tables/table_synth_reg.tex}}

\end{table}

\begin{table}[!ht]\centering\small
\caption{Speedup over a full retrain per method across MedMNIST (unlearning step only).}\label{tab:eff}
\input{tables/table2_efficiency.tex}

\end{table}

\paragraph{Cost.} Unlearning is cheap. Every method except history-replaying FedEraser runs one to two orders of magnitude faster than a full retrain (\textbf{Table~\ref{tab:eff}}), so the dominant cost is the shared pre-training, not the unlearning step.

\paragraph{Where the Field Should Go Next.} \textsc{Lethe}'s findings point to four priorities as below:
\begin{itemize}\itemsep1.5pt\topsep2pt
  \item \textbf{Evaluate on hard requests.} Easy client-level removals leave every method indistinguishable, so stress benchmarks and new methods on class-level and sole-class forgetting, the only regime where they measurably separate.
  \item \textbf{Make membership the target.} On the many clinical tasks that generalize across sites, forgetting a client barely moves the task metric, so the only thing left to erase is residual membership, not forget accuracy.
  \item \textbf{Route requests by hardness.} Since difficulty governs separability, an unlearning system should estimate deletion hardness from class support, sole-provider status, and site skew, and escalate from a cheap update to full retraining only when the request is hard.
  \item \textbf{Design models for deletion.} Because post-hoc weight surgery leaves gentle methods indistinguishable and ascent destructive, the field should move toward deletion-ready designs, such as a shared backbone with site-scoped adapters droppable or retrained on request~\citep{chen2024personalized}.
\end{itemize}

\section{Conclusion}
We propose \textsc{Lethe}, a benchmark for federated unlearning in medical imaging, with two findings. First, forgetting difficulty, not the choice of method, dominates current results, so a discriminative evaluation requires hard-to-forget settings. Second, on tasks that generalize across sites, the erasable signal is membership rather than task contribution.

\bibliographystyle{unsrtnat}
\bibliography{references}

\clearpage
\appendix
\begingroup
\parindent0pt\parskip2.5pt
{\centering\textbf{\large Appendix}\par}
\vspace{8pt}
\textbf{\ref{app:methods}\quad Methods and Hyperparameters}\dotfill\textbf{\pageref{app:methods}}\par
\hspace*{1.4em}\ref{app:methodlist}\quad Unlearning Methods\dotfill\pageref{app:methodlist}\par
\hspace*{1.4em}\ref{app:metrics}\quad Evaluation Metrics\dotfill\pageref{app:metrics}\par
\hspace*{1.4em}\ref{app:taskspec}\quad Architectures and Training\dotfill\pageref{app:taskspec}\par
\textbf{\ref{app:extended}\quad Extended Analyses}\dotfill\textbf{\pageref{app:extended}}\par
\hspace*{1.4em}\ref{app:hetero}\quad Varying Heterogeneity\dotfill\pageref{app:hetero}\par
\hspace*{1.4em}\ref{app:ranking}\quad Method Ranking\dotfill\pageref{app:ranking}\par
\hspace*{1.4em}\ref{app:repr}\quad Class-Level Forgetting and Representations\dotfill\pageref{app:repr}\par
\hspace*{1.4em}\ref{app:scale}\quad Scale and Durability\dotfill\pageref{app:scale}\par
\textbf{\ref{app:privrobust}\quad Privacy, Fidelity, and Robustness}\dotfill\textbf{\pageref{app:privrobust}}\par
\hspace*{1.4em}\ref{app:privacy}\quad Calibrated Membership Inference\dotfill\pageref{app:privacy}\par
\hspace*{1.4em}\ref{app:retraingap}\quad Distance to the Gold Retrain\dotfill\pageref{app:retraingap}\par
\hspace*{1.4em}\ref{app:robustness}\quad Architecture and Sample-Level Robustness\dotfill\pageref{app:robustness}\par
\textbf{\ref{app:vqa}\quad Medical VQA: Granularity and Attack}\dotfill\textbf{\pageref{app:vqa}}\par
\textbf{\ref{app:casestudies}\quad Qualitative Case Studies}\dotfill\textbf{\pageref{app:casestudies}}\par
\endgroup
\vspace{12pt}

\section{Methods and Hyperparameters}\label{app:methods}
\textbf{Table~\ref{tab:main}} lists the twelve unlearning methods evaluated in \textsc{Lethe} alongside the gold-standard \emph{Retrain}. We summarize each procedure and the hyperparameters we use below. Unless stated otherwise, every method starts from the shared pre-trained model, unlearns for five local epochs at the pre-training learning rate, and is then given five FedAvg recovery rounds on the retained clients. Method-specific values follow each method's own paper.

\paragraph{Code and Data Availability.} The repository linked on the first page holds the full benchmark implementation, covering the federated unlearning code, every dataset generator, the medical backbones, and one runnable script per task family. It is released under the Apache License 2.0. Every dataset we use is publicly available and none is redistributed there, so each remains under its provider's original license. The repository \texttt{README} lists the official source of each dataset together with the generator command that builds its federated partition.

\subsection{Unlearning Methods}\label{app:methodlist}
\begin{itemize}\itemsep2pt
  \item \textbf{Retrain}$^\ast$ (gold standard). Retrains from scratch on the retained clients only, the exact model every other method approximates, and has no unlearning hyperparameters.
  \item \textbf{GA} (gradient ascent). The simplest baseline: it ascends the forget loss with the local optimizer, with no method-specific hyperparameters beyond the shared unlearning schedule.
  \item \textbf{PGA}~\citep{halimi}. Projected gradient ascent that maximizes the forget client's loss while projecting each step onto an $L_2$ ball around a reference model averaged from the retained clients, stopping early when validation accuracy falls. \textsc{Lethe} uses an $L_2$ radius $\rho{=}10$.
  \item \textbf{FedEraser}~\citep{federaser}. Reconstructs the model from stored per-round client updates, calibrating each retained update to keep its original magnitude but the newly recomputed direction (calibration ratio $0.5$). It requires the stored update history, so it is omitted from the task families whose history we do not store.
  \item \textbf{Prune}~\citep{classprune}. Scores channels by a per-class TF-IDF of their activations and prunes those most discriminative of the forget class before fine-tuning. Applied at the class level, pruning the top $\rho{=}0.1$ of representation dimensions.
  \item \textbf{PUF}~\citep{puf}. Reuses the standard FedAvg client updates as pseudo-gradients and applies the negated, scaled update of the forget client to the global model, then resumes training to recover.
  \item \textbf{NoT}~\citep{not}. Negates the weights of the first layer of the global model to break inter-layer co-adaptation, then fine-tunes on the retained clients.
  \item \textbf{FedQUIT}~\citep{fedquit}. On-device distillation against a virtual teacher, the global model with the forget sample's true-class output suppressed, matched by a KL objective. \textsc{Lethe} distills at temperature $T{=}4$.
  \item \textbf{FedCARE}~\citep{fedcare}. Conflict-aware projected ascent that maximizes the forget loss while projecting the step onto a half-space preserving a reference loss, paired with a relearning-resistant recovery.
  \item \textbf{FedRecovery}~\citep{fedrecovery}. Subtracts a weighted sum of stored gradient residuals from the converged model and adds calibrated Gaussian noise for differential-privacy indistinguishability from a retrain. \textsc{Lethe} scales the noise to each tensor's standard deviation ($\sigma{=}0.01$).
  \item \textbf{BFU}~\citep{bfu}. Casts the local model as a variational posterior and unlearns by subtracting the forget-data likelihood while a parameter-self-sharing term preserves retained accuracy. \textsc{Lethe} uses a variational anchor $\beta{=}0.05$.
  \item \textbf{FedDNI}~\citep{feddni}. Injects diffusive, structured noise into the forget target to steer the model away from the memorized samples, then restores the global model. \textsc{Lethe} scales the noise to each tensor's standard deviation ($\sigma{=}0.1$).
  \item \textbf{FUSED}~\citep{fused}. Identifies the layers most sensitive to the forget knowledge and trains sparse adapters on the retained data to overwrite it, reversibly.
\end{itemize}

\subsection{Evaluation Metrics}\label{app:metrics}
We formalize every metric reported in the main text. Let $M_0$, $M_u$, and $M^{\ast}$ denote the pre-unlearn, unlearned, and gold-retrain models, and let $D_f$, $D_r$, and $D_t$ be the forget, retain, and test sets.

\paragraph{Utility.} For classification and closed-set VQA we report top-1 accuracy,
\begin{equation}
A(M,D)=\frac{1}{|D|}\sum_{(x,y)\in D}\mathbb{1}\!\left[\arg\max M(x)=y\right],
\end{equation}
as forget $A(M_u,D_f)$, retain $A(M_u,D_r)$, and test $A(M_u,D_t)$ accuracy. Open-ended VQA reports token accuracy, the fraction of generated answer tokens that match the reference answer.

\paragraph{Dense Prediction.} Segmentation uses the Dice overlap between a predicted mask $\hat{Y}$ and the ground truth $Y$,
\begin{equation}
\mathrm{Dice}(\hat{Y},Y)=\frac{2\,|\hat{Y}\cap Y|}{|\hat{Y}|+|Y|}.
\end{equation}
Denoising, cross-modality synthesis, and registration use PSNR and SSIM between a prediction $\hat{x}$ and target $x$,
\begin{equation}
\begin{aligned}
\mathrm{PSNR}(\hat{x},x)&=10\log_{10}\frac{\mathrm{MAX}^2}{\tfrac{1}{n}\sum_{i}(\hat{x}_i-x_i)^2},\\[2pt]
\mathrm{SSIM}(\hat{x},x)&=\frac{(2\mu_{\hat{x}}\mu_x+c_1)(2\sigma_{\hat{x}x}+c_2)}{(\mu_{\hat{x}}^2+\mu_x^2+c_1)(\sigma_{\hat{x}}^2+\sigma_x^2+c_2)},
\end{aligned}
\end{equation}
where $\mathrm{MAX}$ is the signal range and $c_1,c_2$ stabilize the ratio. Lesion localization uses hit-rate, the fraction of predicted lesion centroids $\hat{c}$ that land within a tolerance $\tau$ of the annotated center $c$,
\begin{equation}
\mathrm{HR}=\frac{1}{|D|}\sum_{(x,c)\in D}\mathbb{1}\!\left[\lVert\hat{c}(x)-c\rVert_2\le\tau\right].
\end{equation}

\paragraph{Privacy.} Membership is scored by a per-sample likelihood-ratio attack. For a target sample $z$, reference models calibrate the score distributions of $z$ when it is in and out of training, $\mathcal{N}(\mu_{\mathrm{in}},\sigma_{\mathrm{in}}^2)$ and $\mathcal{N}(\mu_{\mathrm{out}},\sigma_{\mathrm{out}}^2)$, and the attack thresholds the ratio
\begin{equation}
\Lambda(z)=\frac{p\!\left(s(M_u,z)\mid \mathrm{in}\right)}{p\!\left(s(M_u,z)\mid \mathrm{out}\right)},
\end{equation}
with $s(\cdot,\cdot)$ the model's confidence on $z$. We report its AUC and its true-positive rate at a fixed low false-positive rate ($\mathrm{TPR@FPR}$). A model that has truly forgotten $z$ matches $M^{\ast}$ and gives AUC near $0.5$.

\paragraph{Robustness to Backdoors.} Backdoor erasure is the attack-success-rate, the fraction of trigger-carrying inputs still routed to the planted target label $y_t$,
\begin{equation}
\mathrm{ASR}=\frac{1}{|D_b|}\sum_{x\in D_b}\mathbb{1}\!\left[\arg\max M_u(x\oplus t)=y_t\right],
\end{equation}
for trigger pattern $t$ and poisoned set $D_b$.

\paragraph{Cost, Closeness, and Durability.} With $T_M$ the time to produce model $M$, cost is the speedup over a full retrain, $s=T_{M^{\ast}}/T_{M_u}$. Raw closeness to the gold is measured in parameter space by $\Delta_2=\lVert M_u-M^{\ast}\rVert_2$ and in output space by $\mathrm{KL}(M_u\,\Vert\,M^{\ast})$ over the softened predictions. Durability re-exposes the unlearned model to the forget data for $k$ fine-tuning epochs and reports the recovered forget accuracy $R_k=A(M_u^{(k)},D_f)$; a durable erasure keeps $R_k$ low.

\subsection{Architectures and Training}\label{app:taskspec}
\textbf{Table~\ref{tab:taskspec}} gives the federated pre-training configuration of each task family, and \textbf{Table~\ref{tab:arch}} the backbone architectures and parameter counts (datasets and metrics are in the \textbf{Table~\ref{tab:tasks}}). One U-Net is shared across the dense-prediction and generation tasks, differing only in the loss, Dice plus cross-entropy for segmentation, mean-squared error for denoising, a peak-weighted MSE heatmap for localization, and an $L_1$-plus-MSE reconstruction for synthesis. The classification robustness study additionally trains a from-scratch ResNet-18. Each round runs two local epochs, and every task then follows the common unlearning schedule of Sec.~\ref{app:methodlist}, five unlearning epochs and five recovery rounds. All experiments run in PyTorch $2.4.1$ on two NVIDIA RTX $5500$ GPUs.
\begin{table}[t]
\centering
\caption{Federated pre-training configuration per task family: client partition and training schedule (rounds, batch size, learning rate), with two local epochs each. Datasets, backbones, and metrics are in \textbf{Table~\ref{tab:tasks}}. $^{\dagger}$Camelyon17 uses $50$ rounds at batch $256$. $^{\ddagger}$Small dual-encoder shown; the VLM variant trains at batch $32$, learning rate $0.005$.}
\label{tab:taskspec}
\input{tables/table_taskspec.tex}

\end{table}
\begin{table}[t]
\centering
\caption{Backbone architectures and parameter counts. One U-Net serves every dense-prediction and generation task. $|V|$ is the answer vocabulary and $L$ the generated answer length. The VLM image branch is ImageNet-pretrained~\citep{imagenet}. $^{\ast}$FL-trainable parameters; the frozen DistilBERT text encoder runs offline, outside federated training, and is not counted.}
\label{tab:arch}
\resizebox{\textwidth}{!}{\input{tables/table_arch.tex}}

\end{table}

\section{Extended Analyses}\label{app:extended}
We report deeper analyses of the classification findings: the full heterogeneity range, a statistical method ranking, the class-level representation view, and scale and durability studies.

\subsection{Varying Heterogeneity}\label{app:hetero}
\begin{figure}[ht]\centering
\resizebox{\textwidth}{!}{\includegraphics{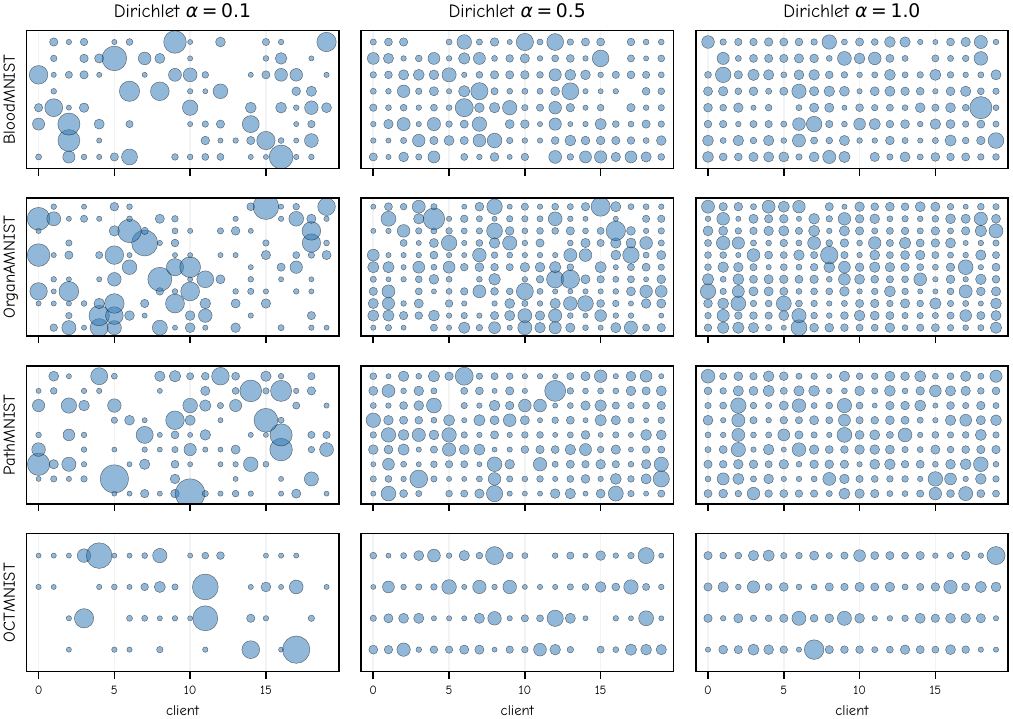}}
\caption{Label heterogeneity of the $20$-client partitions across MedMNIST (rows) at Dirichlet $\alpha\in\{0.1,0.5,1.0\}$ (columns). Each bubble is a (client, class) cell sized by sample count.}\label{fig:bubble}
\end{figure}

The main text summarizes the Dirichlet label skew with one BloodMNIST curve (\textbf{Fig.~\ref{fig:results}(c)}). \textbf{Tables~\ref{tab:alphablood}--\ref{tab:alphaoct}} give it in full: forget (F) and retain (R) accuracy for all twelve methods at Dirichlet $\alpha\in\{0.1,0.5,1.0\}$ across MedMNIST, under client-level forgetting over three seeds. The pattern from the main text holds at every $\alpha$ and on every dataset. The gentle methods stay tightly clustered, while gradient ascent and, at the most heterogeneous settings, FedCARE decline. \textbf{Fig.~\ref{fig:bubble}} shows these partitions.

\begin{table}[htbp]\centering\small
\caption{Varying heterogeneity on BloodMNIST: forget (F) and retain (R) accuracy for all twelve methods at Dirichlet $\alpha\in\{0.1,0.5,1.0\}$ under client-level forgetting.}\label{tab:alphablood}
\input{tables/table_alpha_blood.tex}

\end{table}

\begin{table}[htbp]\centering\small
\caption{Varying heterogeneity on OrganAMNIST: forget (F) and retain (R) accuracy for all twelve methods at Dirichlet $\alpha\in\{0.1,0.5,1.0\}$ under client-level forgetting.}\label{tab:alphaorgana}
\input{tables/table_alpha_organa.tex}

\end{table}

\begin{table}[htbp]\centering\small
\caption{Varying heterogeneity on PathMNIST: forget (F) and retain (R) accuracy for all twelve methods at Dirichlet $\alpha\in\{0.1,0.5,1.0\}$ under client-level forgetting.}\label{tab:alphapath}
\input{tables/table_alpha_path.tex}

\end{table}

\begin{table}[htbp]\centering\small
\caption{Varying heterogeneity on OCTMNIST: forget (F) and retain (R) accuracy for all twelve methods at Dirichlet $\alpha\in\{0.1,0.5,1.0\}$ under client-level forgetting.}\label{tab:alphaoct}
\input{tables/table_alpha_oct.tex}

\end{table}

\subsection{Method Ranking}\label{app:ranking}
\begin{figure}[htbp]\centering
\resizebox{\textwidth}{!}{\includegraphics{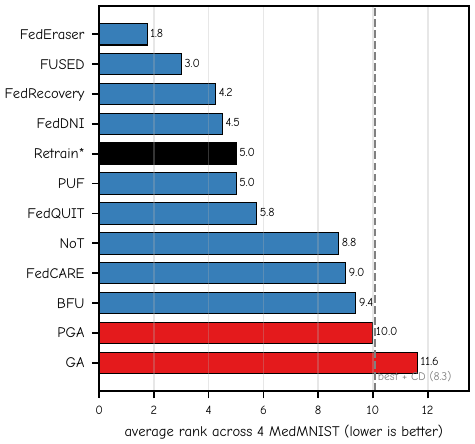}}
\caption{Average rank by retain accuracy across the four MedMNIST client-level datasets (lower is better), with the Nemenyi critical-difference threshold (dashed, best rank $+$ CD).}\label{fig:cd}
\end{figure}

\textbf{Fig.~\ref{fig:cd}} ranks all twelve methods by retain accuracy across the four MedMNIST client-level datasets (Friedman--Nemenyi~\citep{demsar}, $p{<}10^{-3}$). The gentle methods form one indistinguishable cluster around the gold retrain, and only the gradient-ascent methods fall significantly below it. The usable conclusion is to avoid ascent, not to select a single best method, since among the gentle methods no choice is statistically determined.

\subsection{Class-Level Forgetting and Representations}\label{app:repr}
\begin{figure}[tp]\centering
\includegraphics[width=\textwidth]{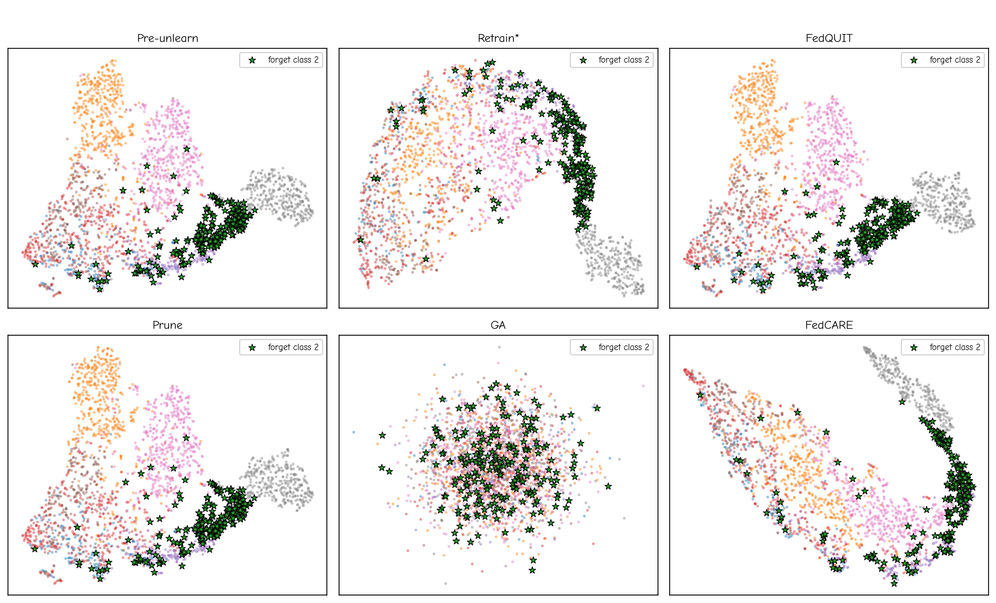}
\caption{Class-level forgetting on BloodMNIST (forget class~$2$, shown as stars): t-SNE~\citep{tsne} of the penultimate features for the pre-unlearn model, the gold retrain, and four representative methods.}\label{fig:tsne}
\end{figure}

Even when a class-level method matches the gold retrain on accuracy, its penultimate features can still cluster the forget class, so the label is erased while the representation is not (\textbf{Fig.~\ref{fig:tsne}}). The sole-class hard-forget protocol (\textbf{Table~\ref{tab:hardforget}}), where the forget client is the only holder of a class, widens the spread among gentle methods.

\begin{table}[htbp]\centering\small
\caption{Hard-forget protocol (the forget client is the sole holder of one class): gentle-method retain accuracy.}\label{tab:hardforget}
\input{tables/table5_hardforget.tex}

\end{table}

\subsection{Scale and Durability}\label{app:scale}
\begin{figure}[t]\centering
\includegraphics[width=\textwidth]{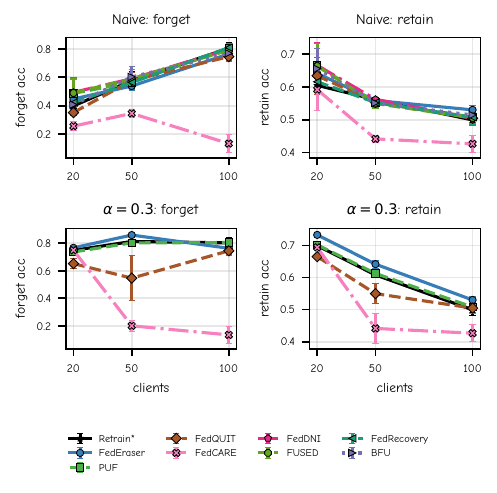}
\caption{Client-count scaling on BloodMNIST. Top row: naive scaling (forget, retain). Bottom row: fixed $\alpha{=}0.3$ (forget, retain).}\label{fig:scalingall}
\end{figure}

\begin{figure}[t]\centering
\resizebox{\textwidth}{!}{\includegraphics{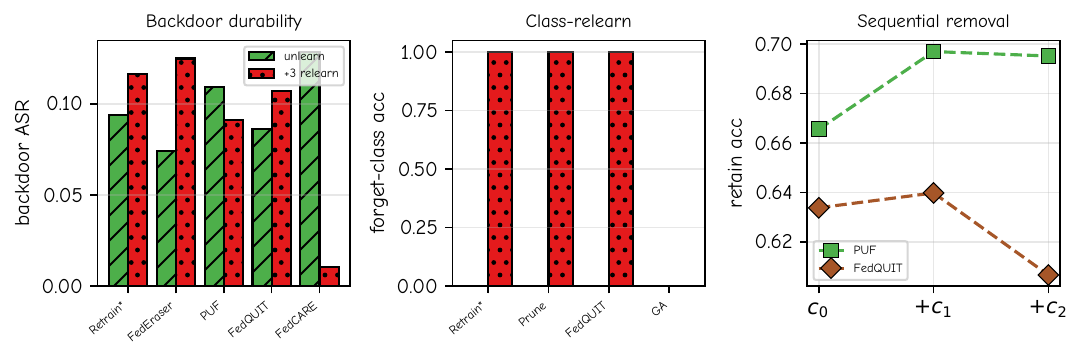}}
\caption{Robustness on BloodMNIST. Left: backdoor attack-success-rate after unlearning and after three relearning epochs. Middle: forget-class accuracy at the same two points. Right: retain accuracy across sequential removals $c_0\to c_1\to c_2$.}\label{fig:robust}
\end{figure}

Two controls separate a real forgetting effect from an artifact: enlarging the federation (\textbf{Fig.~\ref{fig:scalingall}}) and re-exposing the model to the forget data (\textbf{Fig.~\ref{fig:robust}}).

\section{Privacy, Fidelity, and Robustness}\label{app:privrobust}
These subsections test whether the core finding survives a stronger privacy attack, a raw distance to the gold retrain in parameter and output space, and changes of architecture and granularity.

\subsection{Calibrated Membership Inference}\label{app:privacy}
The population membership attack stays near chance, so we add a retrain-calibrated multi-reference U-LiRA on BloodMNIST and VQA-RAD (\textbf{Table~\ref{tab:ulira}}), which exposes residual membership at a low false-positive rate rather than by AUC.

\begin{table}[t]\centering\small
\caption{Multi-reference U-LiRA on BloodMNIST and VQA-RAD (eight OUT reference models): AUC and TPR@$1\%$FPR, chance AUC $0.5$. Dashes mark unavailable runs.}\label{tab:ulira}
\input{tables/table_ulira.tex}

\end{table}

\subsection{Distance to the Gold Retrain}\label{app:retraingap}
The utility and privacy panels read each metric relative to the gold retrain. We also measure raw closeness to it, in parameter space ($\Delta_2$, the $L_2$ weight distance) and in output space (KL of the softened predictions), averaged over each task family (\textbf{Table~\ref{tab:retraingap}}). In raw parameter and output space, the pre-unlearn model is already close to gold ($\Delta_2\approx0.35$ on classification), and the gentle methods merely stay there ($\Delta_2\approx0.2$--$0.5$ at KL$\approx0.01$--$0.08$), while ascent and heavy-noise methods drift away, NoT to $\Delta_2\approx6.6$ and GA to $28$ on VQA. On tasks with little to forget, retrain-gap does not separate the gentle methods, and its main use is to identify the ones that diverge.

\begin{table}[t]\centering\small
\caption{Retrain gap, the parameter-space $L_2$ ($\Delta_2$) and output-space KL between each unlearned model and the same-seed gold retrain, averaged within each task family (lower is closer).}\label{tab:retraingap}
\input{tables/table_retraingap.tex}

\end{table}

\subsection{Architecture and Sample-Level Robustness}\label{app:robustness}
The main-text findings are not an artifact of one architecture or granularity: they persist under sample-level forgetting (\textbf{Table~\ref{tab:sample}}), on a from-scratch ResNet-18 (\textbf{Table~\ref{tab:resnet}}), and for segmentation (\textbf{Table~\ref{tab:segsample}}). \textbf{Table~\ref{tab:coverage}} records which methods run in which scenario.

\begin{table}[tbp]\centering\small
\caption{Sample-level forgetting of a random $10\%$ of forget client~$0$'s samples, with mean forget (F) and retain (R).}\label{tab:sample}
\resizebox{\textwidth}{!}{\input{tables/table9_sample.tex}}

\end{table}

\begin{table}[tbp]\centering\small
\caption{ResNet-18 architecture, client-level forgetting, with forget (F) and retain (R) accuracy across MedMNIST.}\label{tab:resnet}
\resizebox{\textwidth}{!}{\input{tables/table11_resnet.tex}}

\end{table}

\begin{table}[tbp]\centering\small
\caption{Sample-level segmentation forgetting, with mean forget (F) and retain (R) Dice per dataset.}\label{tab:segsample}
\input{tables/table15_segmentation_sample.tex}

\end{table}

\begin{table}[tbp]\centering\small
\caption{Method $\times$ scenario coverage ($\bullet$ = evaluated). Columns: Client, Class, Sample, Multi-client, Sequential, Scaling, Hard-forget, calibrated MIA, ResNet-18, $\alpha$-levels, Backdoor durability. The 2023--2026 additions (FedRecovery, BFU, FedDNI, FUSED) run at client level and on segmentation (\textbf{Tables~\ref{tab:main},~\ref{tab:seg}}).}\label{tab:coverage}
\begin{tabular}{lccccccccccc}
\toprule
Method & Cl & Cls & Smp & MC & Seq & Scl & HF & MIA & RN & $\alpha$ & BD \\
\midrule
Retrain* & $\bullet$ & $\bullet$ & $\bullet$ & $\bullet$ & & $\bullet$ & $\bullet$ & & $\bullet$ & & $\bullet$ \\
GA & $\bullet$ & $\bullet$ & $\bullet$ & & & & & & & & \\
PGA & $\bullet$ & & & & & & & & & & \\
FedEraser & $\bullet$ & & & & & $\bullet$ & $\bullet$ & $\bullet$ & $\bullet$ & $\bullet$ & $\bullet$ \\
Prune & & $\bullet$ & & & & & & & & & \\
PUF & $\bullet$ & & & $\bullet$ & $\bullet$ & $\bullet$ & $\bullet$ & $\bullet$ & $\bullet$ & $\bullet$ & $\bullet$ \\
NoT & $\bullet$ & & & & & & & & & & \\
FedQUIT & $\bullet$ & $\bullet$ & $\bullet$ & $\bullet$ & $\bullet$ & $\bullet$ & $\bullet$ & $\bullet$ & $\bullet$ & $\bullet$ & $\bullet$ \\
FedCARE & $\bullet$ & $\bullet$ & $\bullet$ & & & $\bullet$ & $\bullet$ & $\bullet$ & & $\bullet$ & $\bullet$ \\
\bottomrule
\end{tabular}

\end{table}

\section{Medical VQA: Granularity and Attack}\label{app:vqa}
\begin{figure}[tp]\centering
\resizebox{\textwidth}{!}{\includegraphics{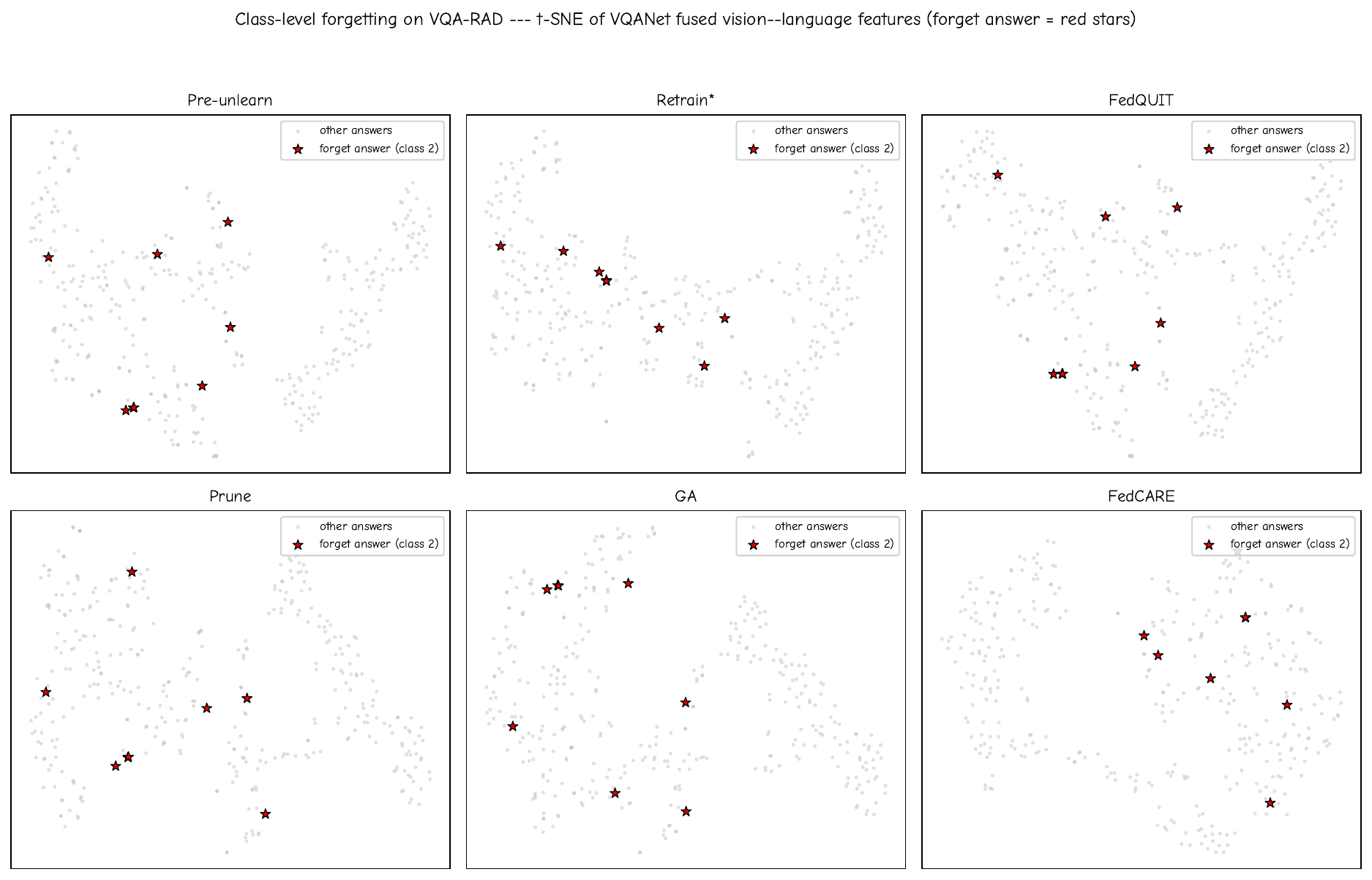}}
\caption{Class-level forgetting on VQA-RAD, t-SNE of the VQANet fused vision-language features (forget answer as red stars).}\label{fig:vqatsne}
\end{figure}

The vision-language finding of the main text holds in two further respects. \textbf{Granularity.} Forgetting a random $10\%$ of QA pairs barely moves either accuracy and leaves MIA near chance, while forgetting an entire answer class drives that answer to $0$ for every method including the gold retrain, the discriminative granularity as in classification (\textbf{Table~\ref{tab:vqagran}}). A t-SNE of the fused features (\textbf{Fig.~\ref{fig:vqatsne}}) shows why the class case is easy. The forget answer's samples are already dispersed among the others, since answers are question-dependent rather than a visual cluster, so little representational structure remains to erase. \textbf{Stronger Attack.} The multi-reference U-LiRA on VQA-RAD (\textbf{Table~\ref{tab:ulira}}) keeps every method's AUC in the near-chance band, with only a mild residual-membership tail.
\begin{table}[tbp]\centering\small
\caption{VQA at finer granularities on the three datasets, with forget (F) and retain (R) answer accuracy and MIA.}\label{tab:vqagran}
\resizebox{\textwidth}{!}{\input{tables/table_vqa_sample.tex}}

\vspace{6pt}

\resizebox{\textwidth}{!}{\input{tables/table_vqa_class.tex}}

\end{table}

\section{Qualitative Case Studies}\label{app:casestudies}
\begin{figure}[tp]\centering
\includegraphics[width=\textwidth]{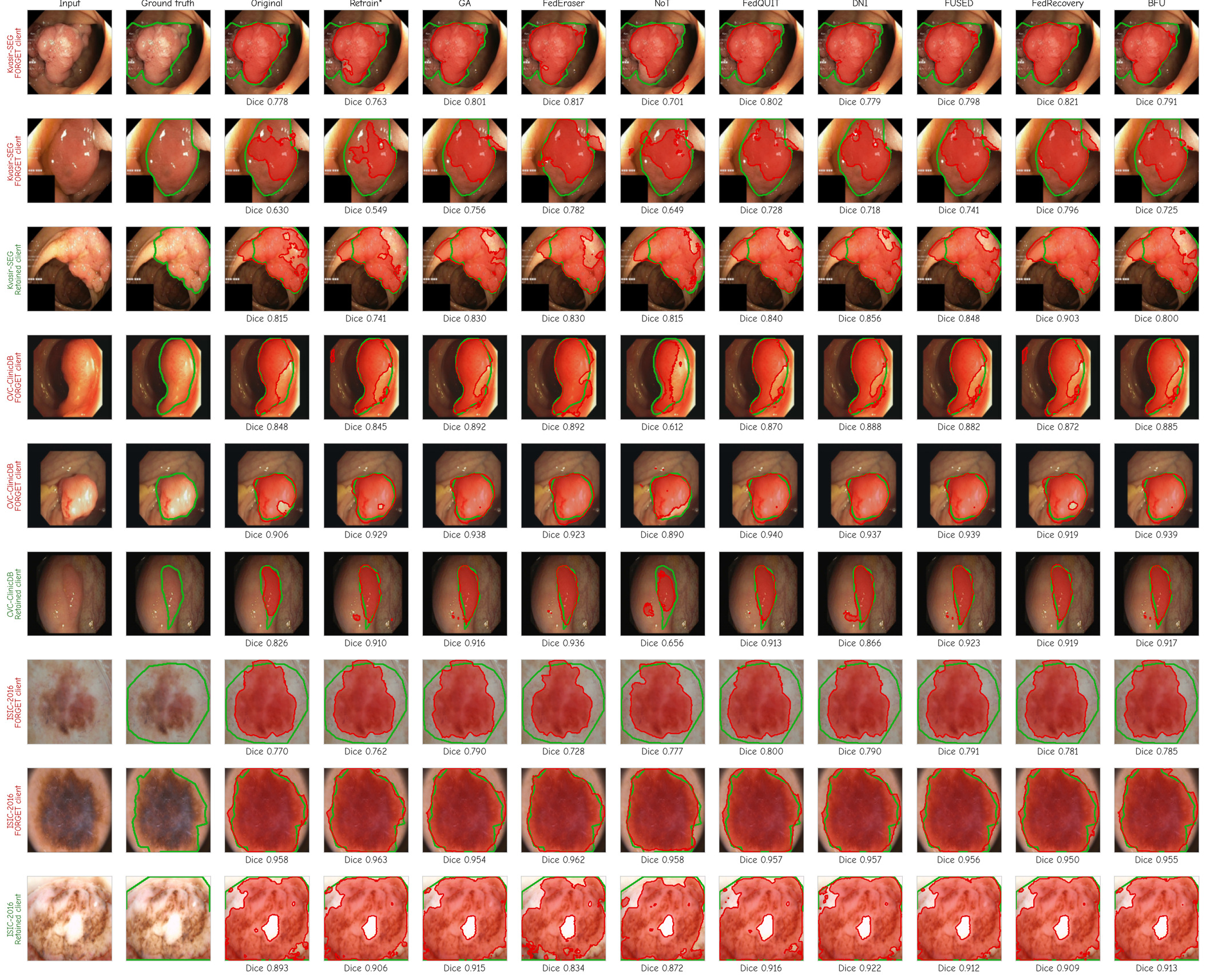}
\caption{Qualitative segmentation case study on the forget client (rows: Kvasir-SEG, CVC-ClinicDB, ISIC-2016). Columns: input, ground truth, pre-unlearn, then each method's prediction with per-image Dice.}\label{fig:segcase}
\end{figure}

\begin{figure}[t]\centering
\includegraphics[width=\textwidth]{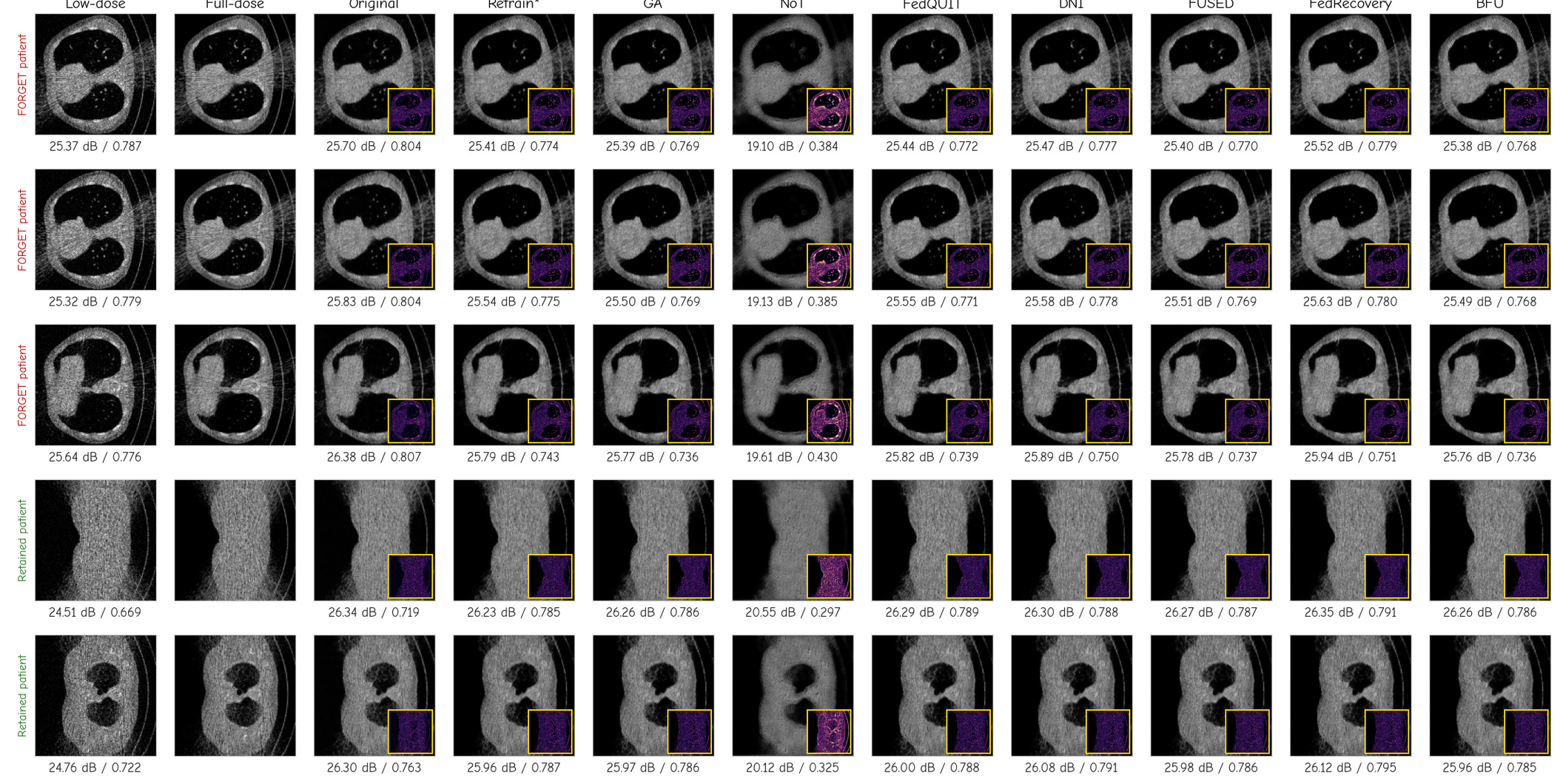}
\caption{Low-dose-CT denoising case study (AAPM-Mayo-2016). Columns: quarter-dose input, full-dose target, pre-unlearn prediction, then each method's restoration of the forget patient.}\label{fig:denoise}
\end{figure}

\begin{figure}[t]\centering
\includegraphics[width=\textwidth]{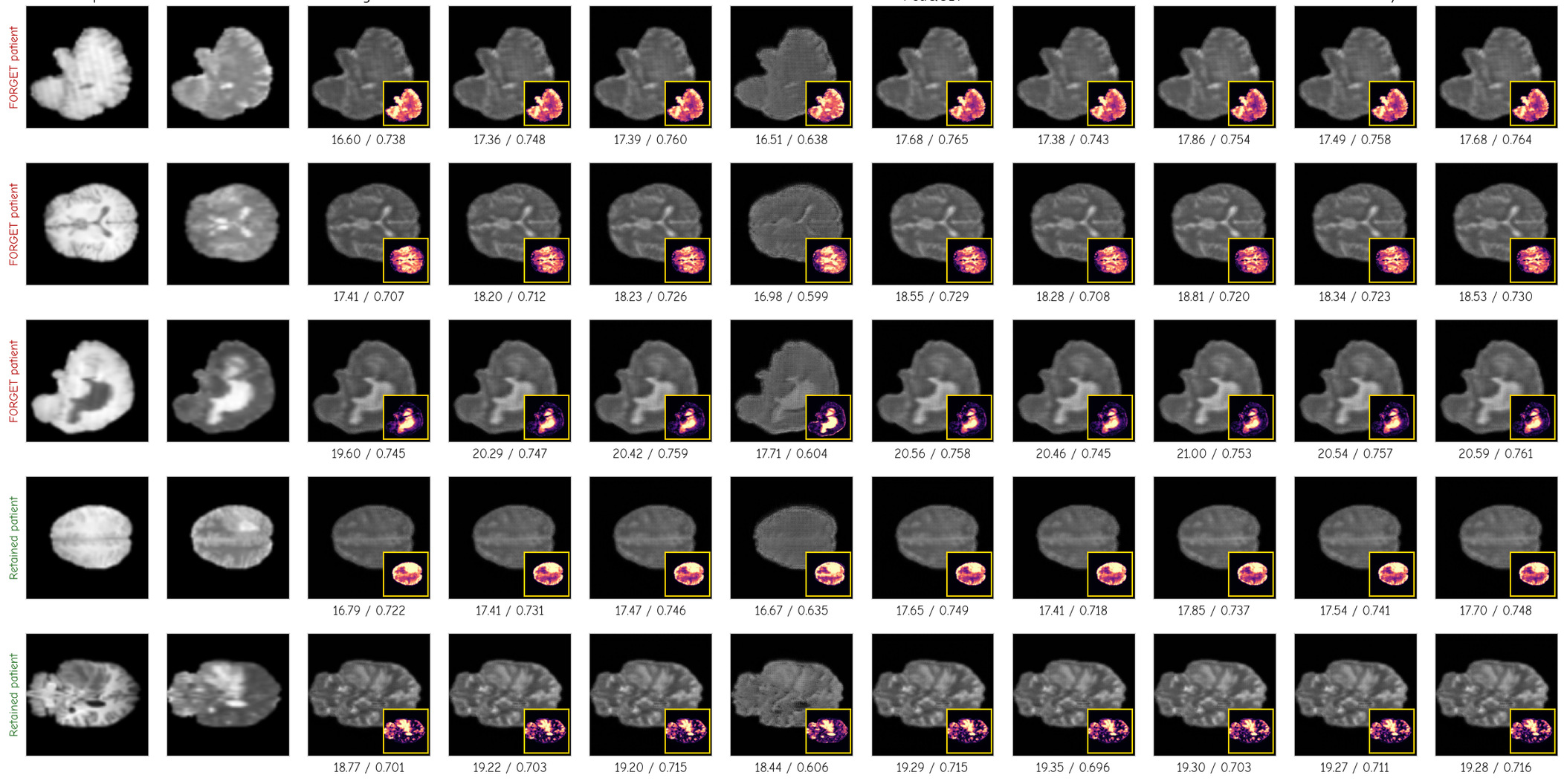}
\caption{Cross-modality synthesis case study (IXI T1$\to$T2): the forget patient's T2 reconstructed by each method, with PSNR / SSIM.}\label{fig:synth}
\end{figure}

\begin{figure}[t]\centering
\includegraphics[width=\textwidth]{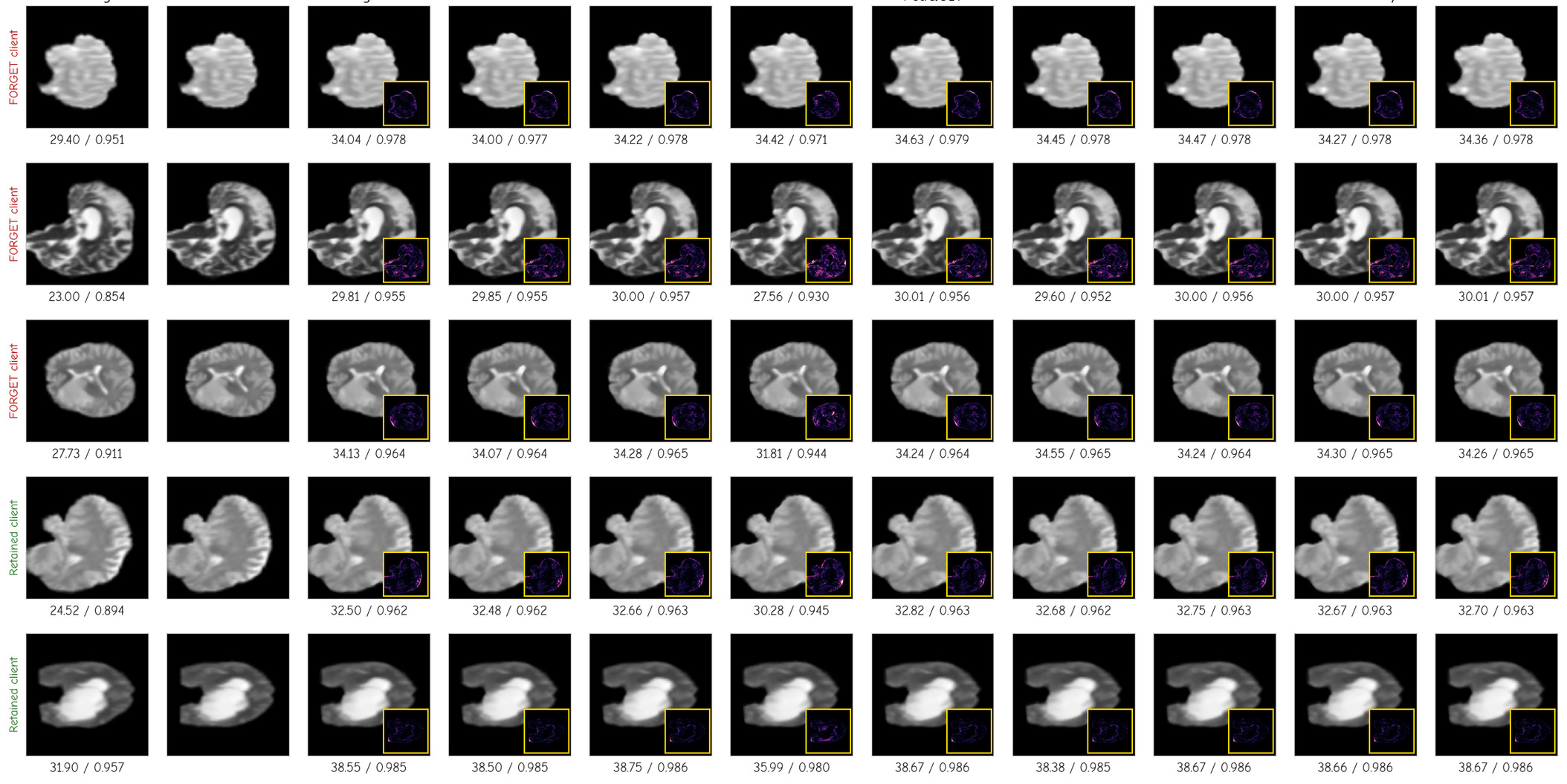}
\caption{Registration case study: each method's warp of the moving image toward the fixed target, with PSNR / SSIM.}\label{fig:reg}
\end{figure}

\begin{figure}[t]\centering
\includegraphics[width=\textwidth]{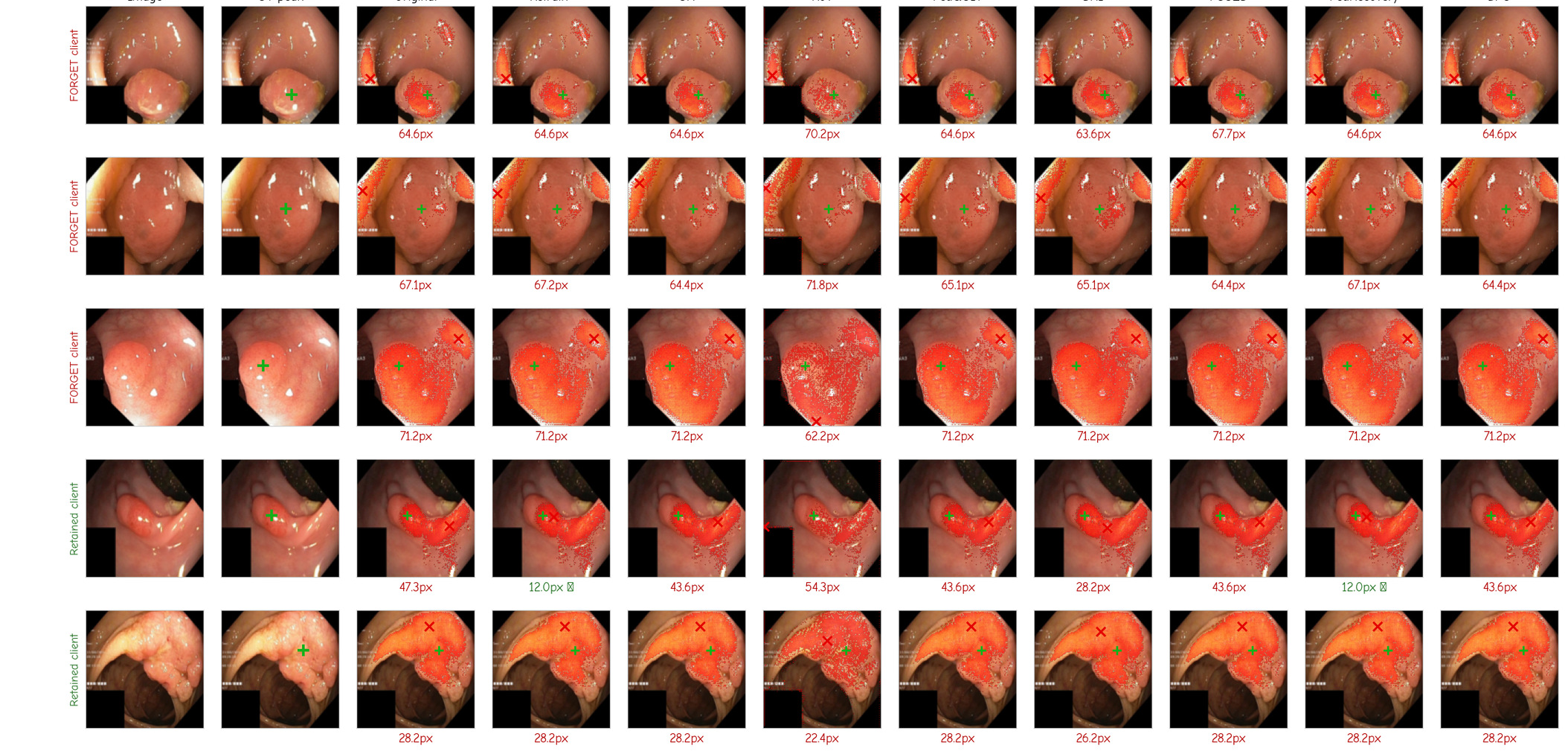}
\caption{Lesion-localization case study: the forget client's predicted lesion centroid per method against the ground truth (green $+$).}\label{fig:detect}
\end{figure}

Per-method qualitative examples on the forget client for each dense-prediction and regression family (\textbf{Figs.~\ref{fig:segcase}--\ref{fig:detect}}). In every case the gold retrain and the gentle methods reconstruct the forget client's data comparably to the retained data, while the collapsing methods degrade visibly.

\end{document}

%% file: tables/table_tasks.tex
\begin{tabular}{@{}llll@{}}
\toprule
Task family & Datasets & Net & Metric \\
\midrule
Classification & 6 datasets & CNN, ResNet & Acc. \\
Segmentation & Kvasir-SEG, CVC, ISIC & U-Net & Dice \\
Denoising & AAPM-Mayo & U-Net & PSNR \\
3D classification & MedMNIST3D & 3D-CNN & Acc. \\
Synthesis & IXI (T1$\to$T2) & U-Net & PSNR/SSIM \\
Registration & IXI & VoxelMorph & PSNR/SSIM \\
Localization & Kvasir polyp & U-Net & Hit-rate \\
VQA & VQA-RAD, SLAKE, PathVQA & Dual-encoder & Ans.\ acc. \\
\bottomrule
\end{tabular}

%% file: tables/table1_main.tex
\begin{tabular}{lcccccccccccc}
\toprule
 & \multicolumn{8}{c}{\textbf{MedMNIST classification}} & \multicolumn{4}{c}{\textbf{Real clinical}} \\
\cmidrule(lr){2-9}\cmidrule(lr){10-13}
Method & \multicolumn{2}{c}{Blood} & \multicolumn{2}{c}{OrganA} & \multicolumn{2}{c}{Path} & \multicolumn{2}{c}{OCT} & \multicolumn{2}{c}{kvasir} & \multicolumn{2}{c}{Camelyon17} \\
\cmidrule(lr){2-3}\cmidrule(lr){4-5}\cmidrule(lr){6-7}\cmidrule(lr){8-9}\cmidrule(lr){10-11}\cmidrule(lr){12-13}
 & F & R & F & R & F & R & F & R & F & R & F & R \\
\midrule
\rowcolor{gray!12}\emph{Full model}$^\dagger$ & 0.485 & 0.642 & 0.148 & 0.760 & 0.054 & 0.563 & 0.233 & 0.518 & -- & -- & -- & -- \\
\rowcolor{gray!12}Retrain$^*$ (gold) & 0.407 & $0.604_{\pm0.012}$ & 0.081 & $0.764_{\pm0.004}$ & 0.023 & $0.553_{\pm0.007}$ & 0.092 & $0.501_{\pm0.029}$ & 0.759 & $0.594_{\pm0.007}$ & 0.784 & $0.687_{\pm0.030}$ \\
\midrule
GA & 0.000 & $0.073_{\pm0.000}$ & 0.000 & $0.061_{\pm0.000}$ & 0.021 & $0.117_{\pm0.000}$ & 1.000 & $0.330_{\pm0.000}$ & 0.000 & $0.134_{\pm0.000}$ & 0.497 & $0.503_{\pm0.000}$ \\
PGA~\citep{halimi} & 0.108 & $0.177_{\pm0.029}$ & 0.022 & $0.195_{\pm0.039}$ & 0.001 & $0.205_{\pm0.008}$ & 0.352 & $0.424_{\pm0.080}$ & 0.000 & $0.133_{\pm0.003}$ & 0.512 & $0.563_{\pm0.045}$ \\
FedEraser~\citep{federaser} & 0.445 & $0.651_{\pm0.027}$ & 0.092 & $0.828_{\pm0.021}$ & 0.022 & $0.617_{\pm0.007}$ & 0.132 & $0.514_{\pm0.024}$ & 0.823 & $0.587_{\pm0.003}$ & 0.685 & $0.711_{\pm0.013}$ \\
PUF~\citep{puf} & 0.462 & $0.641_{\pm0.062}$ & 0.082 & $0.759_{\pm0.006}$ & 0.023 & $0.569_{\pm0.004}$ & 0.050 & $0.492_{\pm0.016}$ & 0.822 & \cellcolor{shadebest}$0.596_{\pm0.005}$& 0.799 & $0.695_{\pm0.024}$ \\
NoT~\citep{not} & 0.183 & $0.217_{\pm0.096}$ & 0.067 & $0.502_{\pm0.009}$ & 0.005 & $0.310_{\pm0.029}$ & 0.000 & $0.479_{\pm0.000}$ & 0.000 & $0.220_{\pm0.055}$ & 0.550 & $0.566_{\pm0.030}$ \\
FedQUIT~\citep{fedquit} & 0.350 & \cellcolor{shadebest}$0.614_{\pm0.051}$& 0.079 & \cellcolor{shadebest}$0.763_{\pm0.005}$& 0.023 & \cellcolor{shade2nd}$0.562_{\pm0.002}$& 0.008 & $0.481_{\pm0.001}$ & 0.000 & $0.501_{\pm0.033}$ & 0.755 & $0.696_{\pm0.012}$ \\
FedCARE~\citep{fedcare} & 0.255 & $0.563_{\pm0.063}$ & 0.052 & $0.557_{\pm0.042}$ & 0.011 & $0.215_{\pm0.061}$ & 0.667 & $0.380_{\pm0.070}$ & 0.400 & $0.126_{\pm0.057}$ & 0.526 & $0.654_{\pm0.036}$ \\
FedRecovery~\citep{fedrecovery} & 0.422 & \cellcolor{shade2nd}$0.615_{\pm0.011}$& 0.082 & $0.760_{\pm0.005}$ & 0.023 & $0.571_{\pm0.002}$ & 0.062 & \cellcolor{shadebest}$0.495_{\pm0.019}$& 0.725 & \cellcolor{shade2nd}$0.591_{\pm0.005}$& 0.744 & \cellcolor{shadebest}$0.692_{\pm0.024}$ \\
BFU~\citep{bfu} & 0.408 & $0.655_{\pm0.060}$ & 0.000 & $0.061_{\pm0.000}$ & 0.021 & $0.117_{\pm0.000}$ & 1.000 & $0.330_{\pm0.000}$ & 0.000 & $0.535_{\pm0.005}$ & 0.497 & $0.503_{\pm0.000}$ \\
FedDNI~\citep{feddni} & 0.494 & $0.665_{\pm0.069}$ & 0.082 & \cellcolor{shade2nd}$0.762_{\pm0.006}$& 0.022 & \cellcolor{shadebest}$0.557_{\pm0.000}$& 0.015 & $0.483_{\pm0.003}$ & 0.706 & $0.607_{\pm0.005}$ & 0.770 & \cellcolor{shade2nd}$0.682_{\pm0.057}$ \\
FUSED~\citep{fused} & 0.489 & $0.663_{\pm0.070}$ & 0.078 & $0.766_{\pm0.005}$ & 0.025 & $0.567_{\pm0.002}$ & 0.052 & \cellcolor{shade2nd}$0.494_{\pm0.003}$& 0.728 & $0.590_{\pm0.004}$ & 0.756 & $0.667_{\pm0.053}$ \\
\bottomrule
\end{tabular}

%% file: tables/table3_classlevel.tex
\begin{tabular}{lccccc}
\toprule
Method (retain$\uparrow$) & blood & organa & path & oct & kvasir \\
\midrule
\rowcolor{gray!12}Retrain* & $0.640_{\pm0.017}$ & $0.719_{\pm0.003}$ & $0.589_{\pm0.004}$ & $0.512_{\pm0.001}$ & $0.636_{\pm0.007}$ \\
GA & $0.078_{\pm0.000}$ & $0.059_{\pm0.000}$ & $0.124_{\pm0.000}$ & $0.373_{\pm0.000}$ & $0.144_{\pm0.000}$ \\
Prune & \cellcolor{shade2nd}$0.650_{\pm0.070}$& \cellcolor{shadebest}$0.719_{\pm0.006}$& \cellcolor{shade2nd}$0.579_{\pm0.012}$& \cellcolor{shade2nd}$0.514_{\pm0.001}$& \cellcolor{shadebest}$0.621_{\pm0.005}$ \\
FedQUIT & \cellcolor{shadebest}$0.649_{\pm0.071}$& \cellcolor{shade2nd}$0.713_{\pm0.002}$& \cellcolor{shadebest}$0.581_{\pm0.013}$& $0.517_{\pm0.004}$ & \cellcolor{shade2nd}$0.573_{\pm0.007}$ \\
FedCARE & $0.273_{\pm0.114}$ & $0.085_{\pm0.061}$ & $0.187_{\pm0.019}$ & \cellcolor{shadebest}$0.511_{\pm0.000}$& $0.138_{\pm0.000}$ \\
\bottomrule
\end{tabular}

%% file: tables/table6_mia.tex
\begin{tabular}{lcccc}
\toprule
calib.\ MIA ($\to .5$) & BloodMNIST & OrganAMNIST & PathMNIST & OCTMNIST \\
\midrule
FedEraser & $0.557_{\pm0.005}$ & $0.497_{\pm0.005}$ & $0.498_{\pm0.002}$ & $0.480_{\pm0.015}$ \\
PUF & $0.549_{\pm0.016}$ & $0.512_{\pm0.001}$ & $0.503_{\pm0.001}$ & $0.502_{\pm0.015}$ \\
FedQUIT & $0.455_{\pm0.019}$ & $0.493_{\pm0.002}$ & $0.501_{\pm0.001}$ & $0.499_{\pm0.009}$ \\
FedCARE & $0.461_{\pm0.002}$ & $0.513_{\pm0.002}$ & $0.499_{\pm0.001}$ & $0.504_{\pm0.010}$ \\
FedDNI & $0.546_{\pm0.016}$ & $0.511_{\pm0.002}$ & $0.503_{\pm0.001}$ & $0.501_{\pm0.015}$ \\
FUSED & $0.516_{\pm0.038}$ & $0.499_{\pm0.001}$ & $0.500_{\pm0.002}$ & $0.490_{\pm0.015}$ \\
FedRecovery & $0.572_{\pm0.003}$ & $0.513_{\pm0.003}$ & $0.503_{\pm0.001}$ & $0.491_{\pm0.006}$ \\
BFU & $0.487_{\pm0.015}$ & -- & -- & -- \\
\bottomrule
\end{tabular}

%% file: tables/table14_segmentation.tex
\begin{tabular}{lccccccccc}
\toprule
Method & \multicolumn{3}{c}{Kvasir-SEG} & \multicolumn{3}{c}{CVC-ClinicDB} & \multicolumn{3}{c}{ISIC-2016} \\
\cmidrule(lr){2-4}\cmidrule(lr){5-7}\cmidrule(lr){8-10}
 & F & R & MIA & F & R & MIA & F & R & MIA \\
\midrule
\textit{Pre-unlearn $F_0$} & \textit{0.570} & \textit{0.640} & -- & \textit{0.700} & \textit{0.660} & -- & \textit{0.860} & \textit{0.890} & -- \\
\midrule
\rowcolor{gray!12}Retrain* & 0.557 & $0.647_{\pm0.017}$ & -- & 0.697 & $0.673_{\pm0.012}$ & -- & 0.864 & $0.891_{\pm0.003}$ & -- \\
GA & 0.551 & $0.648_{\pm0.019}$ & 0.769 & 0.695 & $0.649_{\pm0.020}$ & 0.716 & 0.862 & $0.885_{\pm0.004}$ & 0.644 \\
PGA & 0.478 & $0.584_{\pm0.020}$ & 0.493 & 0.669 & $0.590_{\pm0.034}$ & 0.375 & 0.791 & $0.816_{\pm0.004}$ & 0.538 \\
FedEraser & 0.597 & $0.670_{\pm0.013}$ & 0.505 & 0.720 & $0.737_{\pm0.008}$ & 0.645 & 0.872 & $0.898_{\pm0.005}$ & 0.501 \\
PUF & 0.548 & $0.629_{\pm0.026}$ & 0.705 & 0.712 & \cellcolor{shade2nd}$0.669_{\pm0.020}$& 0.681 & 0.863 & \cellcolor{shadebest}$0.887_{\pm0.006}$& 0.661 \\
NoT & 0.443 & $0.510_{\pm0.015}$ & 0.525 & 0.593 & $0.416_{\pm0.029}$ & 0.420 & 0.795 & $0.847_{\pm0.007}$ & 0.521 \\
FedQUIT & 0.561 & $0.659_{\pm0.011}$ & 0.608 & 0.695 & $0.647_{\pm0.021}$ & 0.659 & 0.853 & $0.878_{\pm0.002}$ & 0.523 \\
FedCARE & 0.386 & $0.306_{\pm0.045}$ & 0.184 & 0.239 & $0.278_{\pm0.004}$ & 0.562 & 0.701 & $0.759_{\pm0.022}$ & 0.569 \\
FedRecovery & 0.564 & $0.652_{\pm0.028}$ & 0.770 & 0.708 & $0.678_{\pm0.016}$ & 0.801 & 0.855 & $0.882_{\pm0.011}$ & 0.636 \\
BFU & 0.554 & \cellcolor{shade2nd}$0.650_{\pm0.023}$& 0.740 & 0.694 & $0.645_{\pm0.021}$ & 0.688 & 0.863 & \cellcolor{shade2nd}$0.886_{\pm0.003}$& 0.629 \\
FedDNI & 0.555 & $0.637_{\pm0.030}$ & 0.730 & 0.712 & \cellcolor{shadebest}$0.671_{\pm0.028}$& 0.677 & 0.856 & $0.880_{\pm0.003}$ & 0.634 \\
FUSED & 0.556 & \cellcolor{shadebest}$0.646_{\pm0.015}$& 0.709 & 0.694 & $0.652_{\pm0.017}$ & 0.723 & 0.856 & $0.882_{\pm0.002}$ & 0.621 \\
\bottomrule
\end{tabular}

%% file: tables/table_vqa_combined.tex
{\setlength{\tabcolsep}{4.5pt}\renewcommand{\arraystretch}{0.98}
\begin{tabular}{lcccc cccc cccc}
\toprule
& \multicolumn{4}{c}{VQA-RAD (radiology)} & \multicolumn{4}{c}{SLAKE (radiology)} & \multicolumn{4}{c}{PathVQA (pathology)} \\
\cmidrule(lr){2-5}\cmidrule(lr){6-9}\cmidrule(lr){10-13}
Method & F & R & MIA & $R_v$ & F & R & MIA & $R_v$ & F & R & MIA & $R_v$ \\
\midrule
\multicolumn{13}{@{}l}{\textit{Closed-set: classification over the 30 most frequent answers}}\\[1pt]
\textit{Pre-unlearn} & \textit{0.49} & \textit{0.47} & -- & \textit{--} & \textit{0.21} & \textit{0.38} & -- & \textit{--} & \textit{0.50} & \textit{0.48} & -- & \textit{--} \\
\rowcolor{gray!12}Retrain* & 0.235 & $0.485_{\pm0.004}$ & 0.515 & 0.57 & 0.164 & $0.365_{\pm0.024}$ & 0.499 & 0.25 & 0.509 & $0.478_{\pm0.024}$ & 0.482 & 0.36 \\
GA & 0.242 & $0.494_{\pm0.014}$ & 0.531 & 0.57 & 0.075 & $0.272_{\pm0.031}$ & 0.469 & 0.25 & 0.564 & $0.512_{\pm0.015}$ & 0.518 & 0.40 \\
PGA & 0.157 & $0.476_{\pm0.000}$ & 0.506 & 0.35 & 0.235 & $0.368_{\pm0.021}$ & 0.492 & 0.31 & 0.322 & $0.392_{\pm0.013}$ & 0.440 & 0.43 \\
FedEraser & 0.157 & $0.478_{\pm0.003}$ & 0.521 & 0.54 & 0.178 & $0.412_{\pm0.046}$ & 0.488 & 0.24 & 0.542 & $0.496_{\pm0.018}$ & 0.501 & 0.37 \\
PUF & 0.222 & $0.478_{\pm0.005}$ & 0.530 & 0.56 & 0.132 & $0.369_{\pm0.035}$ & 0.504 & 0.24 & 0.502 & \cellcolor{shadebest}$0.479_{\pm0.025}$ & 0.497 & 0.38 \\
NoT & 0.157 & $0.476_{\pm0.000}$ & 0.511 & 0.56 & 0.211 & $0.402_{\pm0.028}$ & 0.500 & 0.29 & 0.509 & $0.498_{\pm0.017}$ & 0.515 & 0.40 \\
FedQUIT & 0.235 & $0.492_{\pm0.011}$ & 0.521 & 0.57 & 0.141 & \cellcolor{shade2nd}$0.367_{\pm0.040}$ & 0.504 & 0.25 & 0.487 & $0.494_{\pm0.018}$ & 0.499 & 0.37 \\
FedCARE & 0.157 & $0.478_{\pm0.003}$ & 0.498 & 0.57 & 0.239 & $0.432_{\pm0.026}$ & 0.499 & 0.23 & 0.491 & $0.489_{\pm0.008}$ & 0.496 & 0.37 \\
FedDNI & 0.209 & \cellcolor{shadebest}$0.484_{\pm0.004}$ & 0.533 & 0.56 & 0.127 & \cellcolor{shade2nd}$0.367_{\pm0.036}$ & 0.505 & 0.23 & 0.491 & $0.481_{\pm0.015}$ & 0.494 & 0.38 \\
FUSED & 0.235 & \cellcolor{shade2nd}$0.486_{\pm0.009}$ & 0.525 & 0.56 & 0.143 & $0.369_{\pm0.044}$ & 0.504 & 0.24 & 0.527 & $0.499_{\pm0.022}$ & 0.504 & 0.37 \\
FedRecovery & 0.203 & $0.482_{\pm0.004}$ & 0.529 & 0.56 & 0.122 & $0.372_{\pm0.033}$ & 0.503 & 0.24 & 0.483 & \cellcolor{shade2nd}$0.480_{\pm0.029}$ & 0.492 & 0.38 \\
BFU & 0.216 & $0.483_{\pm0.003}$ & 0.524 & 0.57 & 0.129 & \cellcolor{shadebest}$0.366_{\pm0.033}$ & 0.503 & 0.24 & 0.498 & $0.482_{\pm0.025}$ & 0.498 & 0.37 \\
\midrule
\multicolumn{13}{@{}l}{\textit{Open-ended: free-form generation scored by token accuracy}}\\[1pt]
\textit{Pre-unlearn} & \textit{0.45} & \textit{0.31} & -- & \textit{--} & \textit{0.14} & \textit{0.33} & -- & \textit{--} & \textit{0.46} & \textit{0.37} & -- & \textit{--} \\
\rowcolor{gray!12}Retrain* & 0.177 & $0.294_{\pm0.003}$ & 0.489 & 0.35 & 0.085 & $0.315_{\pm0.005}$ & 0.474 & 0.16 & 0.461 & $0.367_{\pm0.013}$ & 0.493 & 0.30 \\
GA & 0.247 & $0.309_{\pm0.011}$ & 0.501 & 0.35 & 0.096 & $0.322_{\pm0.005}$ & 0.476 & 0.00 & 0.461 & $0.371_{\pm0.010}$ & 0.501 & 0.31 \\
PGA & 0.177 & $0.294_{\pm0.000}$ & 0.458 & 0.08 & 0.080 & $0.264_{\pm0.021}$ & 0.452 & 0.21 & 0.422 & $0.358_{\pm0.010}$ & 0.494 & 0.33 \\
FedEraser & 0.177 & \cellcolor{shadebest}$0.293_{\pm0.001}$ & 0.480 & 0.34 & 0.085 & \cellcolor{shade2nd}$0.316_{\pm0.010}$ & 0.474 & 0.18 & 0.456 & \cellcolor{shadebest}$0.367_{\pm0.014}$ & 0.489 & 0.31 \\
PUF & 0.194 & $0.298_{\pm0.002}$ & 0.499 & 0.34 & 0.085 & $0.318_{\pm0.006}$ & 0.477 & 0.19 & 0.456 & $0.372_{\pm0.006}$ & 0.500 & 0.31 \\
NoT & 0.183 & $0.296_{\pm0.002}$ & 0.523 & 0.34 & 0.108 & $0.316_{\pm0.006}$ & 0.473 & 0.24 & 0.490 & $0.372_{\pm0.008}$ & 0.478 & 0.33 \\
FedQUIT & 0.226 & $0.305_{\pm0.008}$ & 0.499 & 0.35 & 0.086 & $0.321_{\pm0.005}$ & 0.476 & 0.18 & 0.461 & \cellcolor{shade2nd}$0.371_{\pm0.010}$ & 0.501 & 0.31 \\
FedCARE & 0.204 & \cellcolor{shade2nd}$0.296_{\pm0.003}$ & 0.478 & 0.36 & 0.093 & $0.313_{\pm0.006}$ & 0.474 & 0.17 & 0.456 & \cellcolor{shadebest}$0.367_{\pm0.015}$ & 0.500 & 0.30 \\
FedDNI & 0.215 & $0.298_{\pm0.002}$ & 0.502 & 0.35 & 0.085 & $0.320_{\pm0.006}$ & 0.477 & 0.19 & 0.436 & \cellcolor{shadebest}$0.367_{\pm0.009}$ & 0.503 & 0.31 \\
FUSED & 0.253 & $0.310_{\pm0.012}$ & 0.501 & 0.35 & 0.083 & $0.322_{\pm0.006}$ & 0.476 & 0.18 & 0.456 & $0.372_{\pm0.010}$ & 0.501 & 0.31 \\
FedRecovery & 0.183 & \cellcolor{shade2nd}$0.296_{\pm0.002}$ & 0.500 & 0.35 & 0.085 & \cellcolor{shadebest}$0.315_{\pm0.008}$ & 0.476 & 0.18 & 0.471 & $0.374_{\pm0.009}$ & 0.500 & 0.31 \\
BFU & 0.226 & $0.308_{\pm0.011}$ & 0.499 & 0.35 & 0.086 & $0.321_{\pm0.005}$ & 0.475 & 0.18 & 0.461 & $0.372_{\pm0.011}$ & 0.501 & 0.31 \\
\bottomrule
\end{tabular}
}

%% file: tables/table_denoise.tex
\begin{tabular}{lcccc}
\toprule
Method  &  Forget PSNR  &  Retain PSNR  &  Test PSNR  &  MIA  \\
\midrule
\textit{Pre-unlearn $F_0$} & \textit{24.30} & \textit{24.82} & \textit{24.77} & -- \\
\midrule
\rowcolor{gray!12}Retrain*  &  24.66  &  $25.19_{\pm0.04}$  &  25.14  &  0.40  \\
GA  &  24.65  &  $25.18_{\pm0.04}$  &  25.12  &  0.40  \\
PGA  &  24.47  &  $24.98_{\pm0.09}$  &  24.93  &  0.39  \\
PUF  &  24.70  &  $25.25_{\pm0.04}$  &  25.19  &  0.40  \\
NoT  &  16.84  &  $17.43_{\pm3.42}$  &  17.37  &  0.38  \\
FedQUIT  &  24.67  &  \cellcolor{shade2nd}$25.20_{\pm0.04}$  &  25.15  &  0.40  \\
FedCARE  &  24.65  &  \cellcolor{shadebest}$25.19_{\pm0.03}$  &  25.14  &  0.40  \\
FedDNI  &  24.64  &  \cellcolor{shadebest}$25.19_{\pm0.03}$  &  25.13  &  0.40  \\
FUSED  &  24.66  &  \cellcolor{shade2nd}$25.18_{\pm0.04}$  &  25.13  &  0.40  \\
FedRecovery  &  24.70  &  $25.25_{\pm0.04}$  &  25.20  &  0.40  \\
BFU  &  24.65  &  \cellcolor{shade2nd}$25.18_{\pm0.04}$  &  25.13  &  0.40  \\
\bottomrule
\end{tabular}

%% file: tables/table_cls3d_detect.tex
\begin{tabular}{lcccccccccc}
\toprule
Method & \multicolumn{4}{c}{3D classification (accuracy)} & \multicolumn{6}{c}{Lesion localization (hit-rate)} \\
\cmidrule(lr){2-5}\cmidrule(lr){6-11}
 & \multicolumn{2}{c}{Nodule3D} & \multicolumn{2}{c}{Organ3D} & \multicolumn{2}{c}{Kvasir} & \multicolumn{2}{c}{CVC} & \multicolumn{2}{c}{ISIC} \\
\cmidrule(lr){2-3}\cmidrule(lr){4-5}\cmidrule(lr){6-7}\cmidrule(lr){8-9}\cmidrule(lr){10-11}
 & F & R & F & R & F & R & F & R & F & R \\
\midrule
\textit{Pre-unlearn} & \textit{1.00} & \textit{0.79} & \textit{0.12} & \textit{0.27} & \textit{0.33} & \textit{0.21} & \textit{0.12} & \textit{0.19} & \textit{0.30} & \textit{0.45} \\
\midrule
\rowcolor{gray!12}Retrain*  &  1.00  &  $0.79_{\pm0.00}$  &  0.00  &  $0.31_{\pm0.01}$  &  0.31  &  $0.22_{\pm0.04}$  &  0.08  &  $0.19_{\pm0.07}$  &  0.33  &  $0.47_{\pm0.02}$  \\
GA  &  1.00  &  $0.79_{\pm0.00}$  &  0.00  &  $0.15_{\pm0.03}$  &  0.28  &  $0.22_{\pm0.02}$  &  0.08  &  $0.20_{\pm0.07}$  &  0.33  &  $0.46_{\pm0.03}$  \\
PGA  &  1.00  &  $0.79_{\pm0.00}$  &  0.00  &  $0.13_{\pm0.02}$  &  0.25  &  $0.22_{\pm0.03}$  &  0.04  &  $0.20_{\pm0.07}$  &  0.30  &  $0.46_{\pm0.03}$  \\
PUF  &  1.00  &  $0.79_{\pm0.00}$ &  0.00  &  \cellcolor{shadebest}$0.15_{\pm0.03}$  &  0.33  &  \cellcolor{shade2nd}$0.23_{\pm0.02}$  &  0.12  &  \cellcolor{shade2nd}$0.21_{\pm0.08}$  &  0.33  &  \cellcolor{shadebest}$0.47_{\pm0.03}$  \\
NoT  &  1.00  &  $0.79_{\pm0.00}$  &  0.00  &  $0.18_{\pm0.06}$  &  0.08  &  $0.16_{\pm0.02}$  &  0.08  &  $0.09_{\pm0.02}$  &  0.24  &  $0.34_{\pm0.03}$  \\
FedQUIT  &  1.00  &  $0.79_{\pm0.00}$ &  0.00  &  \cellcolor{shadebest}$0.15_{\pm0.03}$  &  0.31  &  \cellcolor{shade2nd}$0.23_{\pm0.02}$  &  0.12  &  \cellcolor{shade2nd}$0.21_{\pm0.07}$  &  0.30  &  \cellcolor{shade2nd}$0.46_{\pm0.02}$  \\
FedCARE  &  1.00  &  $0.79_{\pm0.00}$  &  0.00  &  \cellcolor{shade2nd}$0.13_{\pm0.02}$  &  0.31  &  \cellcolor{shade2nd}$0.21_{\pm0.03}$  &  0.04  &  \cellcolor{shadebest}$0.18_{\pm0.06}$  &  0.30  &  $0.45_{\pm0.03}$  \\
FedDNI  &  1.00  &  $0.79_{\pm0.00}$  &  0.00  &  \cellcolor{shadebest}$0.15_{\pm0.03}$  &  0.28  &  $0.20_{\pm0.03}$  &  0.00  &  \cellcolor{shade2nd}$0.21_{\pm0.02}$  &  0.21  &  \cellcolor{shadebest}$0.47_{\pm0.03}$  \\
FUSED  &  1.00  &  $0.79_{\pm0.00}$  &  0.00  &  \cellcolor{shadebest}$0.15_{\pm0.04}$  &  0.31  &  \cellcolor{shadebest}$0.22_{\pm0.04}$  &  0.12  &  \cellcolor{shadebest}$0.20_{\pm0.07}$  &  0.33  &  \cellcolor{shade2nd}$0.46_{\pm0.02}$  \\
FedRecovery  &  1.00  &  $0.79_{\pm0.00}$  &  0.00  &  \cellcolor{shadebest}$0.15_{\pm0.03}$  &  0.33  &  \cellcolor{shade2nd}$0.23_{\pm0.02}$  &  0.08  &  \cellcolor{shade2nd}$0.21_{\pm0.06}$  &  0.27  &  \cellcolor{shade2nd}$0.48_{\pm0.01}$  \\
BFU  &  1.00  &  $0.79_{\pm0.00}$  &  0.00  &  \cellcolor{shadebest}$0.15_{\pm0.03}$  &  0.28  &  \cellcolor{shadebest}$0.22_{\pm0.02}$  &  0.08  &  \cellcolor{shadebest}$0.20_{\pm0.07}$  &  0.33  &  \cellcolor{shade2nd}$0.46_{\pm0.03}$  \\
\bottomrule
\end{tabular}

%% file: tables/table_synth_reg.tex
\begin{tabular}{lcccccccc}
\toprule
Method & \multicolumn{4}{c}{Synthesis (IXI T1$\to$T2)} & \multicolumn{4}{c}{Registration (VoxelMorph)} \\
\cmidrule(lr){2-5}\cmidrule(lr){6-9}
 & F-PSNR & F-SSIM & R-PSNR & R-SSIM & F-PSNR & F-SSIM & R-PSNR & R-SSIM \\
\midrule
\textit{Pre-unlearn} & \textit{22.54} & \textit{0.762} & \textit{22.83} & \textit{0.758} & \textit{34.34} & \textit{0.971} & \textit{34.74} & \textit{0.971} \\
\midrule
\rowcolor{gray!12}Retrain*  &  22.77  &  0.762  &  $23.01_{\pm0.06}$  &  0.757  &  34.32  &  0.971  &  $34.73_{\pm0.28}$  &  0.971  \\
GA  &  22.82  &  0.778  &  $23.05_{\pm0.06}$  &  0.773  &  34.54  &  0.972  &  $34.93_{\pm0.27}$  &  0.972  \\
PGA  &  22.63  &  0.737  &  $22.81_{\pm0.17}$  &  0.732  &  34.44  &  0.972  &  $34.84_{\pm0.28}$  &  0.972  \\
PUF  &  22.84  &  0.774  &  $23.06_{\pm0.05}$  &  0.769  &  34.53  &  0.972  &  $34.93_{\pm0.27}$  &  \cellcolor{shade2nd}0.972  \\
NoT  &  21.16  &  0.649  &  $21.49_{\pm0.03}$  &  0.650  &  32.66  &  0.960  &  $33.00_{\pm0.28}$  &  0.961  \\
FedQUIT  &  22.85  &  0.780  &  $23.05_{\pm0.05}$  &  0.775  &  34.55  &  0.972  &  $34.95_{\pm0.28}$  &  \cellcolor{shade2nd}0.972  \\
FedCARE  &  22.76  &  0.781  &  \cellcolor{shade2nd}$23.00_{\pm0.05}$  &  0.777  &  34.35  &  0.971  &  \cellcolor{shade2nd}$34.75_{\pm0.27}$  &  \cellcolor{shadebest}0.971  \\
FedDNI  &  22.78  &  0.756  &  \cellcolor{shadebest}$23.01_{\pm0.05}$  &  \cellcolor{shade2nd}0.751  &  34.36  &  0.971  &  \cellcolor{shadebest}$34.73_{\pm0.30}$  &  \cellcolor{shadebest}0.971  \\
FUSED  &  22.91  &  0.765  &  $23.08_{\pm0.04}$  &  \cellcolor{shadebest}0.760  &  34.54  &  0.972  &  $34.93_{\pm0.27}$  &  \cellcolor{shade2nd}0.972  \\
FedRecovery  &  22.84  &  0.773  &  $23.06_{\pm0.05}$  &  0.768  &  34.52  &  0.972  &  $34.87_{\pm0.33}$  &  \cellcolor{shade2nd}0.972  \\
BFU  &  22.86  &  0.779  &  $23.07_{\pm0.06}$  &  0.774  &  34.54  &  0.972  &  $34.93_{\pm0.27}$  &  \cellcolor{shade2nd}0.972  \\
\bottomrule
\end{tabular}

%% file: tables/table2_efficiency.tex
\begin{tabular}{lcccc}
\toprule
Speedup$\times$ vs retrain  &  BloodMNIST  &  OrganAMNIST  &  PathMNIST  &  OCTMNIST  \\
\midrule
\rowcolor{gray!12}Retrain*  &  1.0  &  1.0  &  1.0  &  1.0  \\
GA  &  21.9  &  22.2  &  23.8  &  25.5  \\
PGA  &  25.1  &  26.4  &  28.3  &  28.7  \\
FedEraser  &  0.6  &  0.7  &  0.9  &  1.1  \\
PUF  &  22.4  &  23.2  &  24.8  &  26.1  \\
NoT  &  23.3  &  26.0  &  27.7  &  27.2  \\
FedQUIT  &  24.5  &  22.8  &  24.1  &  23.4  \\
FedCARE  &  30.4  &  30.1  &  33.2  &  29.5  \\
FedDNI  &  \cellcolor{shadebest}107.2  &  \cellcolor{shade2nd}85.7  &  \cellcolor{shadebest}79.5  &  \cellcolor{shadebest}98.5  \\
FUSED  &  \cellcolor{shade2nd}100.9  &  \cellcolor{shadebest}91.5  &  \cellcolor{shade2nd}78.1  &  \cellcolor{shade2nd}89.0  \\
FedRecovery  &  14.9  &  39.0  &  38.6  &  30.3  \\
BFU  &  23.5  &  25.2  &  50.9  &  64.5  \\
\bottomrule
\end{tabular}

%% file: tables/table_taskspec.tex
\setlength{\tabcolsep}{9pt}\renewcommand{\arraystretch}{1.3}
\small
\begin{tabular}{@{}l l ccc@{}}
\toprule
\textbf{Task family} & \textbf{Client partition} & \textbf{Rounds} & \textbf{Batch} & \textbf{LR} \\
\midrule
Classification (2D)$^{\dagger}$ & $20$ (Dirichlet) & $150$ & $64$ & $0.005$ \\
3D classification & $20$ (Dirichlet) & $60$ & $32$ & $0.005$ \\
\addlinespace[2.5pt]
Segmentation & natural (volume skew) & $80$ & $8$ & $0.02$ \\
Denoising & $10$ patients & $80$ & $8$ & $0.005$ \\
Lesion localization & natural & $80$ & $8$ & $0.005$ \\
\addlinespace[2.5pt]
MRI synthesis & $3$ hospitals & $80$ & $8$ & $0.005$ \\
Registration & $3$ hospitals & $80$ & $8$ & $0.02$ \\
\addlinespace[2.5pt]
Medical VQA$^{\ddagger}$ & $8$--$12$ per dataset & $150$ & $64$ & $0.01$ \\
\bottomrule
\end{tabular}

%% file: tables/table_arch.tex
\setlength{\tabcolsep}{5pt}\renewcommand{\arraystretch}{1.3}
\footnotesize
\begin{tabular}{@{}>{\raggedright\arraybackslash}p{2.5cm} >{\raggedright\arraybackslash}p{2.15cm} >{\raggedright\arraybackslash}p{8.6cm} r@{}}
\toprule
\textbf{Backbone} & \textbf{Used for} & \textbf{Structure (channels\,/\,layers)} & \textbf{\#Params} \\
\midrule
Two-block CNN (FedAvgCNN) & Classification (2D) & Conv $3{\to}32$ ($5{\times}5$)+pool, Conv $32{\to}64$ ($5{\times}5$)+pool, FC $1600{\to}512$, linear & $0.88$M \\
3D-CNN (CNN3D) & 3D classification & $3\times$ Conv3d $1{\to}32{\to}64{\to}128$ ($3^3$)+BN+ReLU+pool, GAP, linear & $0.28$M \\
U-Net ($c{=}32$) & Segmentation, denoising, synthesis, localization & Encoder--decoder with skips, base width $32$ ($32{\to}64{\to}128{\to}256$), $1$-channel sigmoid output & $1.93$M \\
RegUNet (VoxelMorph-style) & Registration & U-Net backbone, $2$-channel displacement field, \texttt{grid\_sample} warp & $1.93$M \\
VQANet (dual-encoder) & VQA (closed-set) & Image: $3\times$ Conv $3{\to}32{\to}64{\to}128$ ($3{\times}3$)+BN+pool, GAP ($128$). Question: Embed ($|V|{=}512$, $64$), mean-pool ($64$). Fuse $192{\to}256$, linear & $0.19$M \\
VQANetVLM & VQA (strong backbone) & ImageNet ResNet-18 (FL-finetuned, $512$) $+$ frozen DistilBERT [CLS] ($768$, offline), fuse $256$, linear & $11.58$M$^{\ast}$ \\
VQAGenNet & VQA (open-ended) & VQANet encoders, Linear to $L{\times}256$ vocabulary & $0.38$M \\
\bottomrule
\end{tabular}

%% file: tables/table_alpha_blood.tex
\begin{tabular}{lcccccc}
\toprule
\multicolumn{7}{c}{\textbf{BloodMNIST}} \\
\midrule
Method & \multicolumn{2}{c}{$\alpha{=}0.1$} & \multicolumn{2}{c}{$\alpha{=}0.5$} & \multicolumn{2}{c}{$\alpha{=}1.0$} \\
\cmidrule(lr){2-3}\cmidrule(lr){4-5}\cmidrule(lr){6-7}
 & F & R & F & R & F & R \\
\midrule
\rowcolor{gray!12}Retrain* & 0.407 & 0.604 & 0.651 & 0.686 & 0.704 & 0.722 \\
GA & 0.000 & 0.073 & 0.075 & 0.069 & 0.065 & 0.071 \\
PGA & 0.108 & 0.177 & 0.266 & 0.229 & 0.389 & 0.178 \\
FedEraser & 0.445 & 0.651 & 0.710 & 0.748 & 0.795 & 0.776 \\
PUF & 0.462 & 0.641 & 0.685 & 0.693 & 0.671 & 0.734 \\
NoT & 0.183 & 0.217 & 0.256 & 0.203 & 0.133 & 0.317 \\
FedQUIT & 0.350 & 0.614 & 0.593 & 0.663 & 0.673 & 0.702 \\
FedCARE & 0.255 & 0.563 & 0.669 & 0.694 & 0.662 & 0.714 \\
FedDNI & 0.494 & 0.665 & 0.665 & 0.688 & 0.725 & 0.721 \\
FUSED & 0.489 & 0.663 & 0.656 & 0.686 & 0.690 & 0.717 \\
FedRecovery & 0.422 & 0.615 & 0.660 & 0.689 & 0.719 & 0.720 \\
BFU & 0.408 & 0.655 & 0.665 & 0.690 & 0.707 & 0.726 \\
\bottomrule
\end{tabular}

%% file: tables/table_alpha_organa.tex
\begin{tabular}{lcccccc}
\toprule
\multicolumn{7}{c}{\textbf{OrganAMNIST}} \\
\midrule
Method & \multicolumn{2}{c}{$\alpha{=}0.1$} & \multicolumn{2}{c}{$\alpha{=}0.5$} & \multicolumn{2}{c}{$\alpha{=}1.0$} \\
\cmidrule(lr){2-3}\cmidrule(lr){4-5}\cmidrule(lr){6-7}
 & F & R & F & R & F & R \\
\midrule
\rowcolor{gray!12}Retrain* & 0.081 & 0.764 & 0.828 & 0.826 & 0.820 & 0.832 \\
GA & 0.000 & 0.061 & 0.000 & 0.059 & 0.000 & 0.062 \\
PGA & 0.022 & 0.195 & 0.248 & 0.219 & 0.049 & 0.104 \\
FedEraser & 0.092 & 0.828 & 0.868 & 0.863 & 0.859 & 0.874 \\
PUF & 0.082 & 0.759 & 0.831 & 0.827 & 0.836 & 0.844 \\
NoT & 0.067 & 0.502 & 0.650 & 0.645 & 0.631 & 0.681 \\
FedQUIT & 0.079 & 0.763 & 0.798 & 0.805 & 0.773 & 0.811 \\
FedCARE & 0.052 & 0.557 & 0.845 & 0.836 & 0.808 & 0.837 \\
FedDNI & 0.082 & 0.762 & 0.837 & 0.828 & 0.828 & 0.841 \\
FUSED & 0.078 & 0.766 & 0.820 & 0.818 & 0.731 & 0.811 \\
FedRecovery & 0.082 & 0.760 & 0.830 & 0.827 & 0.836 & 0.844 \\
BFU & 0.000 & 0.061 & 0.000 & 0.059 & 0.000 & 0.062 \\
\bottomrule
\end{tabular}

%% file: tables/table_alpha_path.tex
\begin{tabular}{lcccccc}
\toprule
\multicolumn{7}{c}{\textbf{PathMNIST}} \\
\midrule
Method & \multicolumn{2}{c}{$\alpha{=}0.1$} & \multicolumn{2}{c}{$\alpha{=}0.5$} & \multicolumn{2}{c}{$\alpha{=}1.0$} \\
\cmidrule(lr){2-3}\cmidrule(lr){4-5}\cmidrule(lr){6-7}
 & F & R & F & R & F & R \\
\midrule
\rowcolor{gray!12}Retrain* & 0.023 & 0.553 & 0.479 & 0.671 & 0.546 & 0.630 \\
GA & 0.021 & 0.117 & 0.135 & 0.108 & 0.079 & 0.111 \\
PGA & 0.001 & 0.205 & 0.174 & 0.331 & 0.304 & 0.343 \\
FedEraser & 0.022 & 0.617 & 0.555 & 0.704 & 0.531 & 0.639 \\
PUF & 0.023 & 0.569 & 0.480 & 0.673 & 0.568 & 0.638 \\
NoT & 0.005 & 0.310 & 0.240 & 0.409 & 0.365 & 0.424 \\
FedQUIT & 0.023 & 0.562 & 0.434 & 0.667 & 0.535 & 0.621 \\
FedCARE & 0.011 & 0.215 & 0.477 & 0.680 & 0.554 & 0.627 \\
FedDNI & 0.022 & 0.557 & 0.495 & 0.675 & 0.569 & 0.635 \\
FUSED & 0.025 & 0.567 & 0.477 & 0.677 & 0.568 & 0.629 \\
FedRecovery & 0.023 & 0.571 & 0.479 & 0.673 & 0.567 & 0.637 \\
BFU & 0.021 & 0.117 & 0.135 & 0.108 & 0.079 & 0.111 \\
\bottomrule
\end{tabular}

%% file: tables/table_alpha_oct.tex
\begin{tabular}{lcccccc}
\toprule
\multicolumn{7}{c}{\textbf{OCTMNIST}} \\
\midrule
Method & \multicolumn{2}{c}{$\alpha{=}0.1$} & \multicolumn{2}{c}{$\alpha{=}0.5$} & \multicolumn{2}{c}{$\alpha{=}1.0$} \\
\cmidrule(lr){2-3}\cmidrule(lr){4-5}\cmidrule(lr){6-7}
 & F & R & F & R & F & R \\
\midrule
\rowcolor{gray!12}Retrain* & 0.092 & 0.501 & 0.625 & 0.763 & 0.855 & 0.791 \\
GA & 1.000 & 0.330 & 0.477 & 0.338 & 0.464 & 0.336 \\
PGA & 0.352 & 0.424 & 0.100 & 0.505 & 0.443 & 0.472 \\
FedEraser & 0.132 & 0.514 & 0.726 & 0.787 & 0.867 & 0.809 \\
PUF & 0.050 & 0.492 & 0.652 & 0.771 & 0.855 & 0.796 \\
NoT & 0.000 & 0.479 & 0.488 & 0.713 & 0.817 & 0.730 \\
FedQUIT & 0.008 & 0.481 & 0.608 & 0.764 & 0.851 & 0.790 \\
FedCARE & 0.667 & 0.380 & 0.475 & 0.739 & 0.848 & 0.790 \\
FedDNI & 0.015 & 0.483 & 0.652 & 0.771 & 0.855 & 0.795 \\
FUSED & 0.052 & 0.494 & 0.638 & 0.769 & 0.855 & 0.792 \\
FedRecovery & 0.062 & 0.495 & 0.641 & 0.769 & 0.855 & 0.791 \\
BFU & 1.000 & 0.330 & 0.477 & 0.338 & 0.464 & 0.336 \\
\bottomrule
\end{tabular}

%% file: tables/table5_hardforget.tex
\begin{tabular}{lccc}
\toprule
Method (retain, sole-class) & blood & organa & path \\
\midrule
\rowcolor{gray!12}Retrain* & 0.674 & 0.760 & 0.547 \\
FedEraser & 0.698 & 0.806 & 0.580 \\
PUF & 0.697 & 0.771 & \cellcolor{shadebest}0.545 \\
FedQUIT & 0.656 & \cellcolor{shadebest}0.767& \cellcolor{shade2nd}0.543 \\
FedCARE & 0.388 & 0.716 & 0.248 \\
FedDNI & \cellcolor{shade2nd}0.692& 0.773 & 0.536 \\
FUSED & \cellcolor{shadebest}0.689& \cellcolor{shade2nd}0.770& 0.533 \\
FedRecovery & 0.701 & 0.771 & \cellcolor{shadebest}0.545 \\
BFU & 0.000 & 0.000 & 0.000 \\
\bottomrule
\end{tabular}

%% file: tables/table_ulira.tex
\begin{tabular}{lcccc}
\toprule
Method & \multicolumn{2}{c}{BloodMNIST} & \multicolumn{2}{c}{VQA-RAD} \\
\cmidrule(lr){2-3}\cmidrule(lr){4-5}
 & AUC & TPR@1\% & AUC & TPR@1\% \\
\midrule
\rowcolor{gray!12}Retrain* & 0.48 & 0.01 & 0.54 & 0.07 \\
GA & -- & -- & 0.56 & 0.05 \\
PGA & 0.53 & 0.00 & 0.45 & 0.00 \\
FedEraser & 0.56 & 0.01 & 0.48 & 0.02 \\
PUF & 0.51 & 0.01 & 0.57 & 0.05 \\
NoT & 0.49 & 0.00 & 0.48 & 0.00 \\
FedQUIT & 0.49 & 0.00 & 0.52 & 0.06 \\
FedCARE & 0.47 & 0.01 & 0.51 & 0.03 \\
FedDNI & 0.51 & 0.01 & 0.57 & 0.09 \\
FUSED & 0.52 & 0.01 & 0.57 & 0.04 \\
FedRecovery & 0.55 & 0.02 & 0.57 & 0.05 \\
BFU & 0.51 & 0.01 & 0.56 & 0.06 \\
\bottomrule
\end{tabular}

%% file: tables/table_retraingap.tex
\begin{tabular}{lcccccc}
\toprule
& \multicolumn{2}{c}{Classification (6 ds)} & \multicolumn{2}{c}{Segmentation (3 ds)} & \multicolumn{2}{c}{VQA (3 ds)} \\
Method & $\Delta_2\!\downarrow$ & KL$\downarrow$ & $\Delta_2\!\downarrow$ & KL$\downarrow$ & $\Delta_2\!\downarrow$ & KL$\downarrow$ \\
\midrule
\textit{Pre-unlearn} & \textit{0.35} & \textit{0.040} & \textit{0.41} & \textit{0.042} & -- & -- \\
\midrule
FedEraser & 0.73$\pm$0.42 & 0.060 & 0.95$\pm$0.14 & 0.070 & 0.17 & 0.258 \\
GA & -- & -- & 0.44$\pm$0.10 & 0.023 & 27.99 & 0.811 \\
PGA & -- & -- & 0.84$\pm$0.35 & 0.114 & 2.71 & 0.890 \\
NoT & -- & -- & 6.63$\pm$0.03 & 0.164 & 6.56 & 0.103 \\
PUF & 0.38$\pm$0.51 & 0.042 & 0.44$\pm$0.10 & 0.022 & 0.44 & 0.010 \\
FedQUIT & 0.48$\pm$0.49 & 0.080 & 0.50$\pm$0.13 & 0.025 & 0.44 & 0.060 \\
FedCARE & 1.50$\pm$0.85 & 1.702 & 0.53$\pm$0.13 & 0.386 & 0.44 & 0.140 \\
FedDNI & 1.56$\pm$0.31 & 0.040 & 2.53$\pm$0.02 & 0.022 & 18.08 & 0.015 \\
FUSED & 0.45$\pm$0.50 & 0.040 & 0.46$\pm$0.11 & 0.023 & 0.47 & 0.051 \\
FedRecovery & 0.31$\pm$0.14 & 0.004 & 0.53$\pm$0.09 & 0.021 & 1.88 & 0.009 \\
BFU & 0.47$\pm$0.69 & 0.106 & 0.43$\pm$0.10 & 0.022 & 0.44 & 0.008 \\
\bottomrule
\end{tabular}

%% file: tables/table9_sample.tex
\begin{tabular}{lcccccccccc}
\toprule
Method & \multicolumn{2}{c}{BloodMNIST} & \multicolumn{2}{c}{OrganAMNIST} & \multicolumn{2}{c}{PathMNIST} & \multicolumn{2}{c}{OCTMNIST} & \multicolumn{2}{c}{kvasir} \\
\cmidrule(lr){2-3}\cmidrule(lr){4-5}\cmidrule(lr){6-7}\cmidrule(lr){8-9}\cmidrule(lr){10-11}
 & F & R & F & R & F & R & F & R & F & R \\
\midrule
\rowcolor{gray!12}Retrain* & 0.381 & $0.597_{\pm0.017}$ & 0.129 & $0.720_{\pm0.005}$ & 0.044 & $0.533_{\pm0.001}$ & 0.164 & $0.509_{\pm0.026}$ & 1.000 & $0.616_{\pm0.005}$ \\
GA & 0.429 & $0.666_{\pm0.065}$ & 0.000 & $0.057_{\pm0.000}$ & 0.019 & $0.111_{\pm0.000}$ & 1.000 & $0.342_{\pm0.000}$ & 0.615 & $0.253_{\pm0.104}$ \\
FedQUIT & 0.452 & \cellcolor{shade2nd}$0.662_{\pm0.069}$& 0.146 & \cellcolor{shadebest}$0.721_{\pm0.005}$& 0.048 & \cellcolor{shadebest}$0.536_{\pm0.002}$& 0.057 & \cellcolor{shadebest}$0.482_{\pm0.012}$& 0.974 & \cellcolor{shade2nd}$0.608_{\pm0.002}$ \\
FedCARE & 0.333 & \cellcolor{shadebest}$0.624_{\pm0.066}$& 0.084 & $0.684_{\pm0.002}$ & 0.019 & \cellcolor{shade2nd}$0.434_{\pm0.016}$ & 0.000 & $0.471_{\pm0.000}$ & 0.000 & $0.529_{\pm0.014}$ \\
BFU & 0.452 & $0.667_{\pm0.067}$ & 0.133 & \cellcolor{shade2nd}$0.719_{\pm0.004}$& 0.046 & \cellcolor{shadebest}$0.536_{\pm0.002}$& 0.019 & \cellcolor{shade2nd}$0.473_{\pm0.001}$& 0.974 & \cellcolor{shadebest}$0.615_{\pm0.004}$ \\
\bottomrule
\end{tabular}

%% file: tables/table11_resnet.tex
\begin{tabular}{lcccccccc}
\toprule
Method & \multicolumn{2}{c}{BloodMNIST} & \multicolumn{2}{c}{OrganAMNIST} & \multicolumn{2}{c}{PathMNIST} & \multicolumn{2}{c}{OCTMNIST} \\
\cmidrule(lr){2-3}\cmidrule(lr){4-5}\cmidrule(lr){6-7}\cmidrule(lr){8-9}
 & F & R & F & R & F & R & F & R \\
\midrule
\rowcolor{gray!12}Retrain* & 0.457 & $0.619_{\pm0.010}$ & 0.321 & $0.853_{\pm0.005}$ & 0.179 & $0.493_{\pm0.019}$ & 0.984 & $0.462_{\pm0.111}$ \\
PUF & 0.515 & \cellcolor{shadebest}$0.625_{\pm0.025}$ & 0.376 & \cellcolor{shade2nd}$0.872_{\pm0.010}$ & 0.099 & $0.482_{\pm0.058}$ & 0.958 & $0.612_{\pm0.078}$ \\
FedQUIT & 0.333 & $0.488_{\pm0.028}$ & 0.048 & $0.712_{\pm0.004}$ & 0.034 & \cellcolor{shade2nd}$0.483_{\pm0.008}$ & 0.748 & \cellcolor{shade2nd}$0.543_{\pm0.049}$ \\
FedDNI & 0.471 & \cellcolor{shade2nd}$0.611_{\pm0.018}$ & 0.323 & \cellcolor{shadebest}$0.858_{\pm0.003}$ & 0.100 & \cellcolor{shadebest}$0.494_{\pm0.043}$ & 0.969 & \cellcolor{shadebest}$0.462_{\pm0.109}$ \\
FUSED & 0.403 & $0.548_{\pm0.018}$ & 0.062 & $0.768_{\pm0.043}$ & 0.026 & $0.479_{\pm0.009}$ & 0.897 & $0.549_{\pm0.124}$ \\
BFU & 0.469 & $0.607_{\pm0.021}$ & 0.060 & $0.780_{\pm0.049}$ & 0.007 & $0.359_{\pm0.062}$ & 0.674 & $0.668_{\pm0.019}$ \\
\bottomrule
\end{tabular}

%% file: tables/table15_segmentation_sample.tex
\begin{tabular}{lcccccc}
\toprule
Method & \multicolumn{2}{c}{Kvasir-SEG} & \multicolumn{2}{c}{CVC-ClinicDB} & \multicolumn{2}{c}{ISIC-2016} \\
\cmidrule(lr){2-3}\cmidrule(lr){4-5}\cmidrule(lr){6-7}
 & F & R & F & R & F & R \\
\midrule
\rowcolor{gray!12}Retrain* & 0.586 & 0.652 & 0.712 & 0.658 & 0.882 & 0.876 \\
GA & 0.541 & 0.625 & 0.681 & 0.672 & 0.891 & 0.883 \\
FedQUIT & 0.536 & \cellcolor{shade2nd}0.616& 0.694 & \cellcolor{shadebest}0.679& 0.896 & \cellcolor{shadebest}0.880 \\
FedCARE & 0.522 & 0.454 & 0.294 & 0.432 & 0.432 & 0.419 \\
BFU & 0.564 & \cellcolor{shadebest}0.656& 0.680 & \cellcolor{shade2nd}0.686& 0.892 & \cellcolor{shade2nd}0.888 \\
\bottomrule
\end{tabular}

%% file: tables/table_vqa_sample.tex
\begin{tabular}{lccccccccc}
\toprule
\multicolumn{10}{c}{\textbf{Sample-level: random $10\%$ of QA pairs}} \\
\midrule
Method & \multicolumn{3}{c}{VQA-RAD (radiology)} & \multicolumn{3}{c}{SLAKE (radiology)} & \multicolumn{3}{c}{PathVQA (pathology)} \\
\cmidrule(lr){2-4}\cmidrule(lr){5-7}\cmidrule(lr){8-10}
 & F & R & MIA & F & R & MIA & F & R & MIA \\
\midrule
\rowcolor{gray!12}Retrain* & 0.467 & $0.476_{\pm0.013}$ & 0.542 & 0.286 & $0.379_{\pm0.039}$ & 0.515 & 0.407 & $0.478_{\pm0.025}$ & 0.535 \\
GA & 0.467 & $0.476_{\pm0.016}$ & 0.551 & 0.286 & $0.398_{\pm0.028}$ & 0.516 & 0.481 & $0.499_{\pm0.016}$ & 0.540 \\
NoT & 0.400 & $0.471_{\pm0.008}$ & 0.538 & 0.333 & $0.425_{\pm0.023}$ & 0.529 & 0.444 & $0.475_{\pm0.015}$ & 0.494 \\
FedQUIT & 0.467 & \cellcolor{shadebest}$0.477_{\pm0.014}$& 0.556 & 0.286 & \cellcolor{shadebest}$0.398_{\pm0.027}$& 0.523 & 0.444 & \cellcolor{shade2nd}$0.500_{\pm0.017}$& 0.543 \\
FedCARE & 0.467 & \cellcolor{shade2nd}$0.468_{\pm0.014}$ & 0.578 & 0.333 & \cellcolor{shade2nd}$0.421_{\pm0.021}$ & 0.499 & 0.370 & \cellcolor{shadebest}$0.481_{\pm0.018}$& 0.481 \\
BFU & 0.467 & \cellcolor{shadebest}$0.477_{\pm0.014}$& 0.556 & 0.309 & \cellcolor{shadebest}$0.398_{\pm0.024}$& 0.525 & 0.407 & $0.505_{\pm0.018}$ & 0.579 \\
\bottomrule
\end{tabular}

%% file: tables/table_vqa_class.tex
\begin{tabular}{lccccccccc}
\toprule
\multicolumn{10}{c}{\textbf{Class-level: forget answer class $2$}} \\
\midrule
Method & \multicolumn{3}{c}{VQA-RAD (radiology)} & \multicolumn{3}{c}{SLAKE (radiology)} & \multicolumn{3}{c}{PathVQA (pathology)} \\
\cmidrule(lr){2-4}\cmidrule(lr){5-7}\cmidrule(lr){8-10}
 & F & R & MIA & F & R & MIA & F & R & MIA \\
\midrule
\rowcolor{gray!12}Retrain* & 0.000 & $0.473_{\pm0.011}$ & 0.554 & 0.000 & $0.410_{\pm0.028}$ & 0.402 & 0.000 & $0.473_{\pm0.030}$ & 0.500 \\
GA & 0.000 & $0.487_{\pm0.009}$ & 0.554 & 0.000 & $0.404_{\pm0.024}$ & 0.433 & 0.000 & $0.507_{\pm0.020}$ & 0.445 \\
NoT & 0.000 & $0.473_{\pm0.009}$ & 0.450 & 0.000 & $0.430_{\pm0.021}$ & 0.477 & 0.059 & $0.480_{\pm0.020}$ & 0.514 \\
FedQUIT & 0.000 & \cellcolor{shadebest}$0.482_{\pm0.003}$& 0.571 & 0.000 & \cellcolor{shade2nd}$0.401_{\pm0.022}$& 0.488 & 0.098 & \cellcolor{shade2nd}$0.506_{\pm0.017}$& 0.508 \\
FedCARE & 0.000 & $0.490_{\pm0.004}$ & 0.671 & 0.000 & $0.423_{\pm0.037}$ & 0.401 & 0.000 & \cellcolor{shade2nd}$0.506_{\pm0.015}$ & 0.499 \\
Prune & 0.000 & $0.485_{\pm0.010}$ & 0.577 & 0.000 & $0.397_{\pm0.029}$ & 0.510 & 0.039 & $0.507_{\pm0.018}$ & 0.511 \\
BFU & 0.000 & \cellcolor{shade2nd}$0.483_{\pm0.009}$& 0.576 & 0.000 & \cellcolor{shadebest}$0.411_{\pm0.022}$& 0.505 & 0.078 & \cellcolor{shadebest}$0.505_{\pm0.017}$& 0.511 \\
\bottomrule
\end{tabular}